\pdfoutput=1

\documentclass[11pt]{article}

\usepackage[]{acl}

\usepackage{times}
\usepackage{amsmath} \usepackage{amssymb}
\usepackage{latexsym}
\usepackage{graphicx}
\usepackage{multirow}
\usepackage{mathtools}
\usepackage{multicol}
\usepackage{array}
\usepackage{enumerate}
\usepackage{booktabs}
\usepackage{amsfonts,amssymb}
\usepackage{graphicx}
\usepackage{multirow}
\usepackage[table,xcdraw]{xcolor}
\usepackage{tabularx}
\usepackage{algorithm}
\usepackage{algpseudocode}
\usepackage{bbding}
\usepackage{makecell}
\usepackage{CJKutf8}
\usepackage[T1]{fontenc}

\usepackage[utf8]{inputenc}

\usepackage{microtype}
\usepackage{amsmath}
\usepackage{autobreak}
\usepackage{pifont}

\usepackage{inconsolata}
\usepackage{threeparttable}
\usepackage{setspace}
\usepackage{subcaption}
\title{PATIMT-Bench: A Multi-Scenario Benchmark for Position-Aware Text Image Machine Translation in Large Vision-Language Models}

\author{WanruZhuang$^{1}$\thanks{~~Equal contribution.}, Wenbo Li$^{1}$\footnotemark[1], Zhibin Lan$^{1}$, Xu Han$^{2}$, Peng Li$^{2}$, \textbf{Jinsong Su\textsuperscript{1,3}\thanks{~~Corresponding author.}} \\
   $^1$School of Informatics, Xiamen University, China \\$^2$Tsinghua, Beijing, China \\ $^3$Shanghai Artificial Intelligence Laboratory, China
    \\
 \texttt{\small \{zhuangwanru, liwenbo\}@stu.xmu.edu.cn}~~
  \texttt{\small jssu@xmu.edu.cn}
 \\}

\begin{document}
\maketitle
\begin{abstract}
 
Text Image Machine Translation (TIMT) aims to translate texts embedded within an image into another language.
Current TIMT studies primarily focus on providing translations for all the text within an image, while neglecting to provide bounding boxes and covering limited scenarios. In this work, we extend traditional TIMT into position-aware TIMT (PATIMT), aiming to support fine-grained and layout-preserving translation, which holds great practical value but remains largely unexplored.
This task comprises two key sub-tasks: region-specific translation and full-image translation with grounding. To support existing models on PATIMT and conduct fair evaluation, we construct the PATIMT benchmark (PATIMT-Bench), which consists of 10 diverse real-world scenarios. Specifically, we introduce an Adaptive Image OCR Refinement Pipeline, which adaptively selects appropriate OCR tools based on scenario and refines the results of text-rich images. To ensure evaluation reliability, we further construct a test set, which contains 1,200 high-quality instances manually annotated and reviewed by human experts.
After fine-tuning on our data, compact Large Vision-Language Models (LVLMs) achieve state-of-the-art performance on both sub-tasks. Experimental results also highlight the scalability and generalizability of our training data\footnote{Our benchmark and code are openly available at https://github.com/XMUDeepLIT/PATIMT-Bench}. 
\end{abstract}

\section{Introduction}
\begin{figure}[t]
    \centering \includegraphics[width=0.48\textwidth]{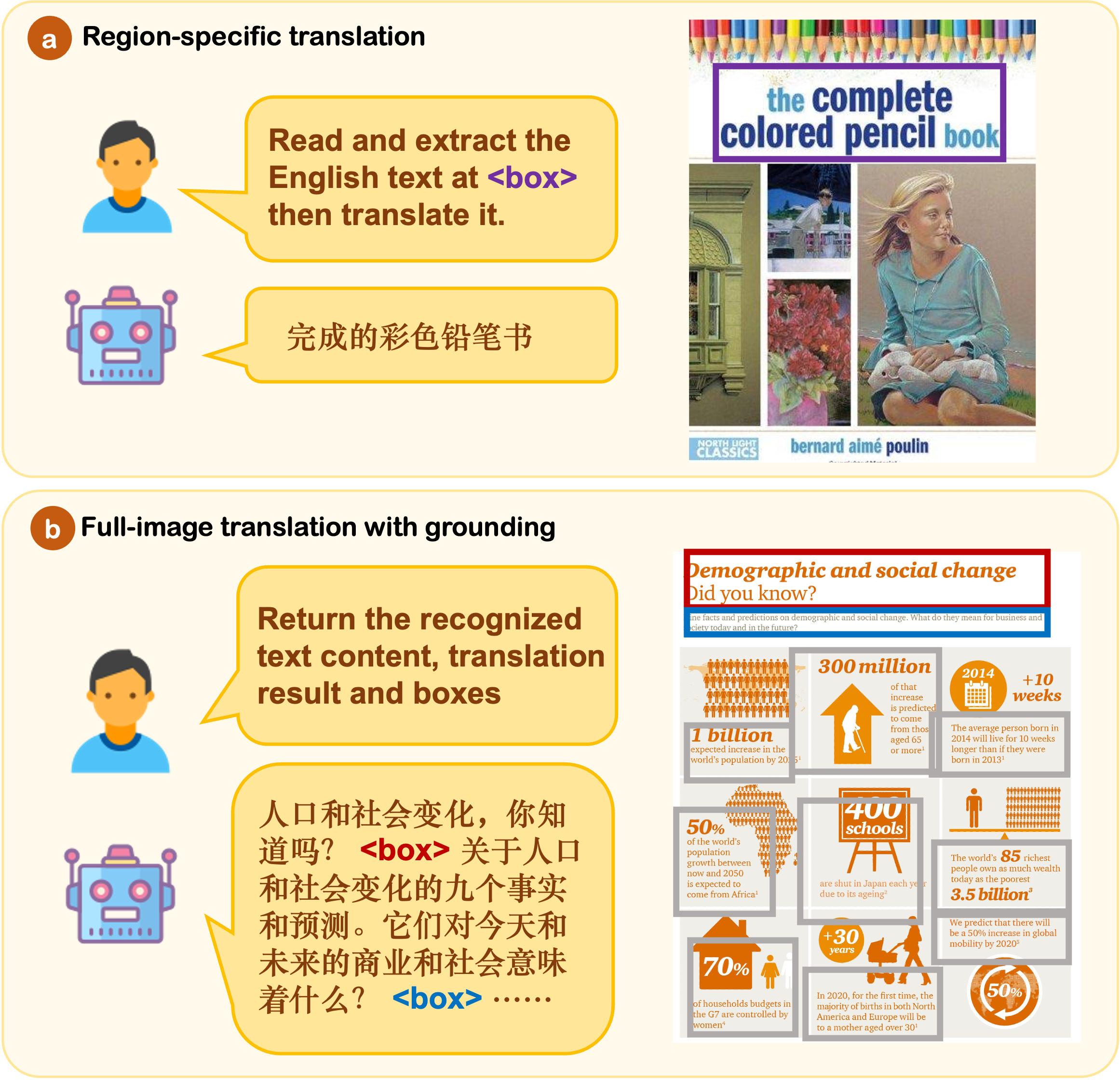} 
    \caption{Two sub-tasks of PATIMT: region-specific translation and full-image translation with grounding.}
    \label{img/intro_task}
\end{figure}
Text Image Machine Translation (TIMT) is a challenging branch of Neural Machine Translation (NMT), offering broad application prospects in both academic research and commercial applications.
Conventional TIMT methods \cite{relatedwork/e2e/50422,relatedwork/e2e/ZhuLLX23,relatedwork/e2e/DIMT} typically focus on generating plain text or markdown-formatted translations of all text within an image, failing to precisely preserve the original layout of source text in the image.
This limitation gives rise to a critical position alignment problem in real-world applications, where users cannot reliably match translations to the corresponding source text. Additionally, all these methods overlook the localized translation requirements.
This significantly limits their practical usability.

In this paper, we explore position-aware TIMT (PATIMT) which contains two core sub-tasks: region-specific translation and full-image translation with grounding as shown in Figure \ref{img/intro_task}.
Region-specific translation enables users to manually select one region of an image for translation, which allows fine-grained, user-controllable TIMT. Full-image translation with grounding ensures precise positional alignment between the translation and source text in the image, enabling seamless rendering of translated image version.

\begin{figure}[t]
    \centering 
    \includegraphics[width=0.49\textwidth]{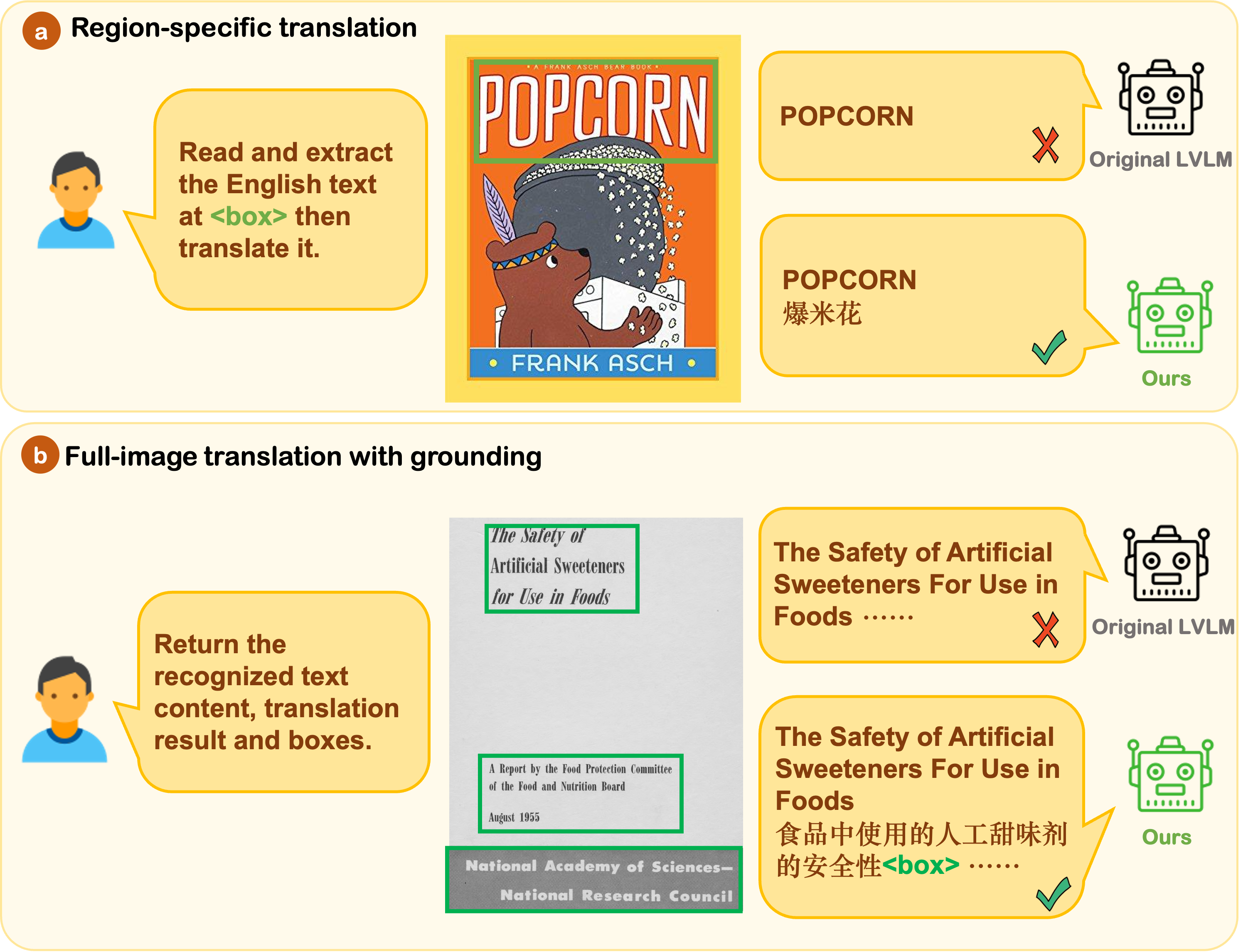} 
    \caption{Comparison between the Original LVLM and the LVLM fine-tuned on our data (Ours). In two types of fine-grained TIMT tasks, ours can correctly follow the translation instructions and conduct precise text referring and grounding within proper layout.  }
    \label{img/intro_comparison}
\end{figure}

Recently, Large Vision-Language Models (LVLMs) \cite{DBLP:journals/corr/abs-2410-18558,DBLP:journals/corr/deepseekvl2,DBLP:journals/corr/internvl2.5,DBLP:journals/corr/qwen2.5-vl} show remarkable performance across diverse multimodal benchmarks such as OCR \cite{OCRBench}, image understanding \cite{DBLP:conf/icdar/ocrvqa,DBLP:conf/wacv/docvqa,DBLP:conf/wacv/InfoVQA,DBLP:conf/acl/chartqa,ScienceQA} and visual grounding \cite{RefCOCO,DBLP:conf/cvpr/refcoco,objdetection_in_wild,countbench}.
They appear to have great potential to perform region-specific translation and full-image translation with grounding. However, existing LVLMs usually fail to follow the above two types of translation instructions, as illustrated in Figure \ref{img/intro_comparison}.
This limitation primarily derives from data scarcity.
Available TIMT datasets \cite{readingbank,relatedwork/cascade/layoutdit,ocrmt30k,DBLP:conf/coling/mit10m} typically lack bounding box annotations or suffer from limited scenarios and scale, making them difficult to support accurate position-aware TIMT. 
Moreover, there is a lack of comprehensive benchmark.
Existing TIMT benchmarks mainly focus on evaluating plain text or markdown translations and only specialize in a single scenario \cite{readingbank,ocrmt30k,relatedwork/e2e/ZhuLLX23,relatedwork/e2e/DIMT}. Although MIT-10M \cite{DBLP:conf/coling/mit10m} covers diverse image categories, it lacks bounding box annotations and is confined to simple scenarios, excluding document, infographic images and so on.
    
Nevertheless, constructing multi-scenario PATIMT datasets remains challenging for three main reasons. 1) General OCR tools typically provide line-by-line recognition results, leading to semantically incoherent annotations; 2) Document-specific OCR tools sometimes ignore text-containing areas and are not always optimal for other scenarios;
 3) Manual annotation is labor-intensive and expensive.
In this work, we address these issues by introducing an automated data processing pipeline to construct a \textbf{high-quality, multi-scenario} PATIMT dataset and a comprehensive benchmark: First, we introduce an \textbf{adaptive image OCR refinement pipeline} that combines a general EasyOCR\footnote{https://github.com/JaidedAI/EasyOCR} with a PDF-optimized MinerU \cite{minerU} to adaptively process images from different scenarios and refine the results of text-rich samples.
Second, we propose \textbf{PATIMT-Bench}, which is explicitly designed to evaluate region-specific translation and full-image translation with grounding for images from diverse domains. Specifically, we use our pipeline to construct training data, which provides fine-grained bounding boxes in proper layout for text within the images.
As for the test set, we select 1200 images with high-quality manual annotations that are carefully reviewed by human experts.

Experimental results demonstrate that all compact LVLMs achieve state-of-the-art performance on PATIMT-Bench after fine-tuning on our training data, outperforming larger models such as Qwen2.5-VL-72B and closed-source models like GPT-4o. A series of systematic analyses are conducted which demonstrate the scalability and generalizability of our dataset.


\section{Related Work}
\subsection{TIMT models}

Early approaches predominantly rely on cascaded systems \cite{relatedwork/cascade/10099625,relatedwork/cascade/layoutdit,ocrmt30k}, where OCR and NMT models are separately optimized and pipelined. Such methods suffer from error propagation, and they only provide plain text translation. Recent advancements explore end-to-end frameworks \cite{relatedwork/e2e/50422,relatedwork/e2e/MaZTHWZ022,relatedwork/e2e/ZhuLLX23,relatedwork/e2e/DIMT} to mitigate these issues.
A representative approach developed by \citet{relatedwork/e2e/DIMT} enables markdown format translations for document-style images.
While this approach achieves layout-aware translation, it remains limitations in handling other scenarios such as infographic, chart, natural scene where markdown is inadequate to establish accurate localization correspondence.

\subsection{TIMT Datasets}
Dataset scarcity remains a critical challenge in TIMT research \cite{shen2024surveymultimodalmachinetranslation}. Early studies \cite{Mansimov2020,relatedwork/e2e/50422,relatedwork/e2e/MaZTHWZ022,NiuM024} primarily rely on synthetic data, which are generated by rendering source language text onto background images.
However, synthetic data are unable to capture the nuanced complexity of text in real-world translation applications (e.g., occlusions, irregular layouts), leading to an inevitable performance gap.

Recent efforts aim to construct real-world TIMT datasets. \citet{ocrmt30k} develops OCRMT30K, derived from street view images and their OCR annotations; \citet{relatedwork/cascade/layoutdit} constructs DITrans, which considers reading order in document images;  \citet{relatedwork/e2e/DIMT} introduces DoTA, a document image machine translation dataset in markdown format. Afterwards, \citet{DBLP:conf/coling/mit10m} constructs MIT-10M, a large-scale, real-world dataset with diverse image categories.
However, it lacks bounding box annotations and  omits more complex scenarios such as infographic and document.
In this work, we propose the Adaptive Image OCR Refinement Pipeline, an automated and cost-effective solution for processing text within images. Our pipeline provides bounding box labels in proper layouts for images from varying scenarios. Table \ref{figures:data_compare} shows the comparison of our dataset with existing TIMT datasets, image examples and output format comparison are listed in Appendix \ref{sec:appendix_data_compare}.

\begin{table}[]
\centering
\begin{scriptsize}
\setlength{\tabcolsep}{1.5mm}
\renewcommand\arraystretch{1.5}
\begin{tabular*}{0.48\textwidth}{l|ccc}
\toprule[1pt]

\textbf{Dataset} & \textbf{Source} & \textbf{Bounding box} & \textbf{Scenario} \\ \hline
\multicolumn{1}{l|}{OCRMT30K \cite{ocrmt30k}} & realistic & \checkmark & street-view \\
\multicolumn{1}{l|}{DITrans \cite{relatedwork/cascade/layoutdit}} & realistic &	\checkmark & document  \\
\multicolumn{1}{l|}{DoTA \cite{relatedwork/e2e/DIMT}} & realistic &	\ding{53} & document  \\
\multicolumn{1}{l|}{UMTIT \cite{NiuM024}} & synthetic &	\ding{53} & document  \\
\multicolumn{1}{l|}{MIT-10M \cite{DBLP:conf/coling/mit10m}} & realistic &	\ding{53} & multi-scene  \\
\multicolumn{1}{l|}{PATIMT-Bench (Ours)} & realistic &	\checkmark & multi-scene  \\

\bottomrule[1pt]
\end{tabular*}
\caption{Comparison of PATIMT-Bench with other TIMT datasets.}
\label{figures:data_compare}
\end{scriptsize}
\end{table}

\subsection{Large Vision-Language Models}

Recent advances in LVLMs \cite{DBLP:journals/corr/abs-2410-18558,DBLP:journals/corr/deepseekvl2,DBLP:journals/corr/internvl2.5,DBLP:journals/corr/qwen2.5-vl} demonstrate remarkable performance across diverse multimodal benchmarks, including visual question answering \cite{DBLP:conf/wacv/docvqa,DBLP:conf/cvpr/textvqa,DBLP:conf/icdar/ocrvqa,DBLP:conf/wacv/InfoVQA,ScienceQA}, OCR \cite{OCRBench,DBLP:journals/corr/OCRBenchv2}, and visual grounding \cite{RefCOCO,DBLP:conf/cvpr/refcoco}. The prevailing architecture integrates a powerful visual encoder with a large language model (LLM) via cross-modal adapters. This unified framework exhibits two strengths: (1) superior translation quality with the powerful LLM, and (2) precise text grounding that enabling position-aware TIMT for diverse images.
Despite these potentials, no existing work has systematically explored position-aware TIMT capability for LVLMs.
In this work, we present PATIMT-Bench, which is designed to evaluate PATIMT through region-specific translation and full-image translation with grounding tasks.


\section{Adaptive Image OCR Refinement Pipeline}
\begin{figure*}[t]
    \centering \includegraphics[width=0.98\textwidth]{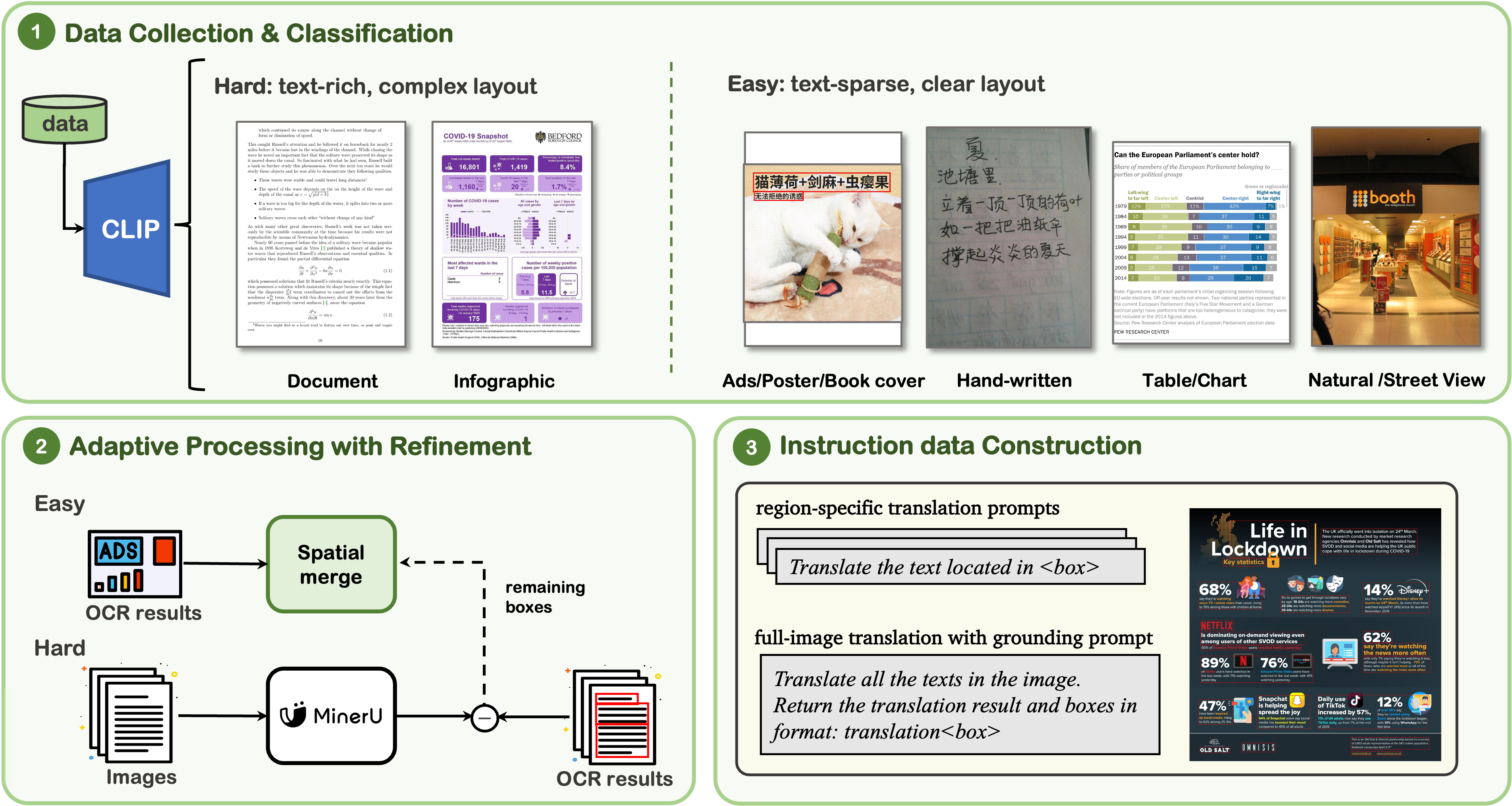} 
    \caption{Our pipeline includes three steps: (1) collecting images from open-source datasets with initial OCR filtering, classifying images into easy/hard categories using CLIP; (2) adaptive processing with refinement strategy that generates accurate annotations for both easy and hard categories, and (3) constructing instruction dataset using GPT-4o. }
    \label{img/pipeline}
\end{figure*}

To develop a high-quality PATIMT dataset for diverse real-world scenarios, we first extensively collect existing open-source image-text datasets and classify these images into corresponding scenarios. Secondly, we introduce an adaptive processing with refinement strategy to adaptively process images from different scenarios. Finally, we prompt GPT-4o \cite{DBLP:journals/corr/abs-2410-21276} to generate the instruction tuning data. Figure \ref{img/pipeline} illustrates the overall pipeline.

\subsection{Data Collection and Preprocessing}

We collect data from the following sources: MIT-10M \cite{DBLP:conf/coling/mit10m}, CC12M \cite{DBLP:conf/cvpr/CC12m}, DocVQA \cite{DBLP:conf/wacv/docvqa}, 
InfoVQA \cite{DBLP:conf/wacv/InfoVQA}, 
TextVQA \cite{DBLP:conf/cvpr/textvqa}, ChartQA \cite{DBLP:conf/acl/chartqa}, Wukong \cite{DBLP:conf/nips/Wukong}, WTW \cite{Long_2021_ICCV/WTW}, LSVT \cite{DBLP:conf/icdar/LSVT}, CDLA\footnote{https://github.com/buptlihang/CDLA.git}, pdfa-eng-wds\footnote{https://huggingface.co/datasets/pixparse/pdfa-eng-wds}, an English hand-wriitten OCR dataset from Nexdata\footnote{from https://www.nexdata.ai}, and some online data.

\begin{figure}[t]
    \centering \includegraphics[width=0.45\textwidth]{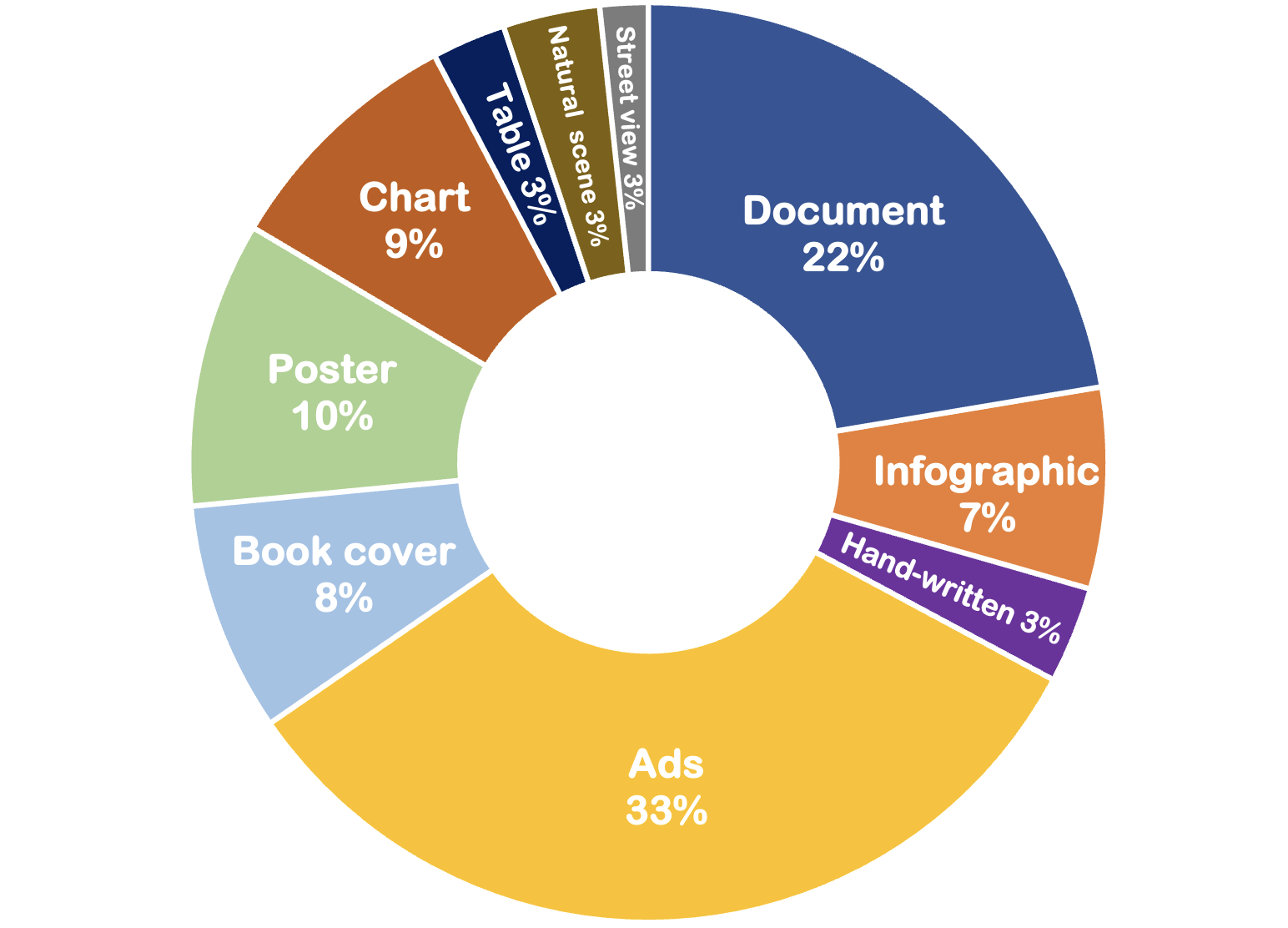} 
    \caption{The proportion of different scenarios in our dataset.}
    \label{img/data}
\end{figure}

After collecting data from various sources, we categorize the collected data using CLIP \cite{DBLP:conf/icml/clip} following \citet{DBLP:journals/corr/abs-2306-17107}, resulting in 10 different scenarios: \textit{advertisement}, \textit{poster}, \textit{book cover}, \textit{natural scene}, \textit{street view}, \textit{chart}, 
\textit{table}, 
\textit{hand-written}, \textit{infographic}, and \textit{document}, Appendix \ref{sec:appendix_clip_cls} lists the detailed implementation. 
To ensure the classification accuracy, we randomly sample 200 images from the easy and hard categories separately for verification, which achieves an accuracy rate of 98.5\%. Figure \ref{img/data} shows the proportion of different scenarios. 

We then implement EasyOCR to generate the OCR results for the collected images and conduct coarse-grained filtering.
Specifically, images are excluded if: (1) their OCR results are empty; (2) their OCR results contain repetitive character sequences of length $\geq$ 3; or (3) the average character pixels are less than 3\% of the total image pixels.

\subsection{Adaptive Processing with Refinement} 
This process handles images from different scenarios and generates box annotations. To begin with, we categorize the aforementioned ten scenarios into two groups based on the level of difficulty:
\vspace{-0.2\topsep}
\begin{itemize}
\setlength{\itemsep}{0.25cm}
\setlength{\parsep}{0.5pt}
\setlength{\parskip}{0.5pt}
\item\textbf{Easy:} Characterized by images containing sparse text, clean layouts, and typically low resolutions with modest aspect ratios.

\item\textbf{Hard:} Characterized by text-rich images with small font sizes, complex layouts, and high resolutions with potentially extreme aspect ratios.
These scenarios present intricate spatial arrangements where text, graphics, and tables are sometimes interleaved.
\end{itemize}

\begin{figure}[t]
    \centering \includegraphics[width=0.48\textwidth]{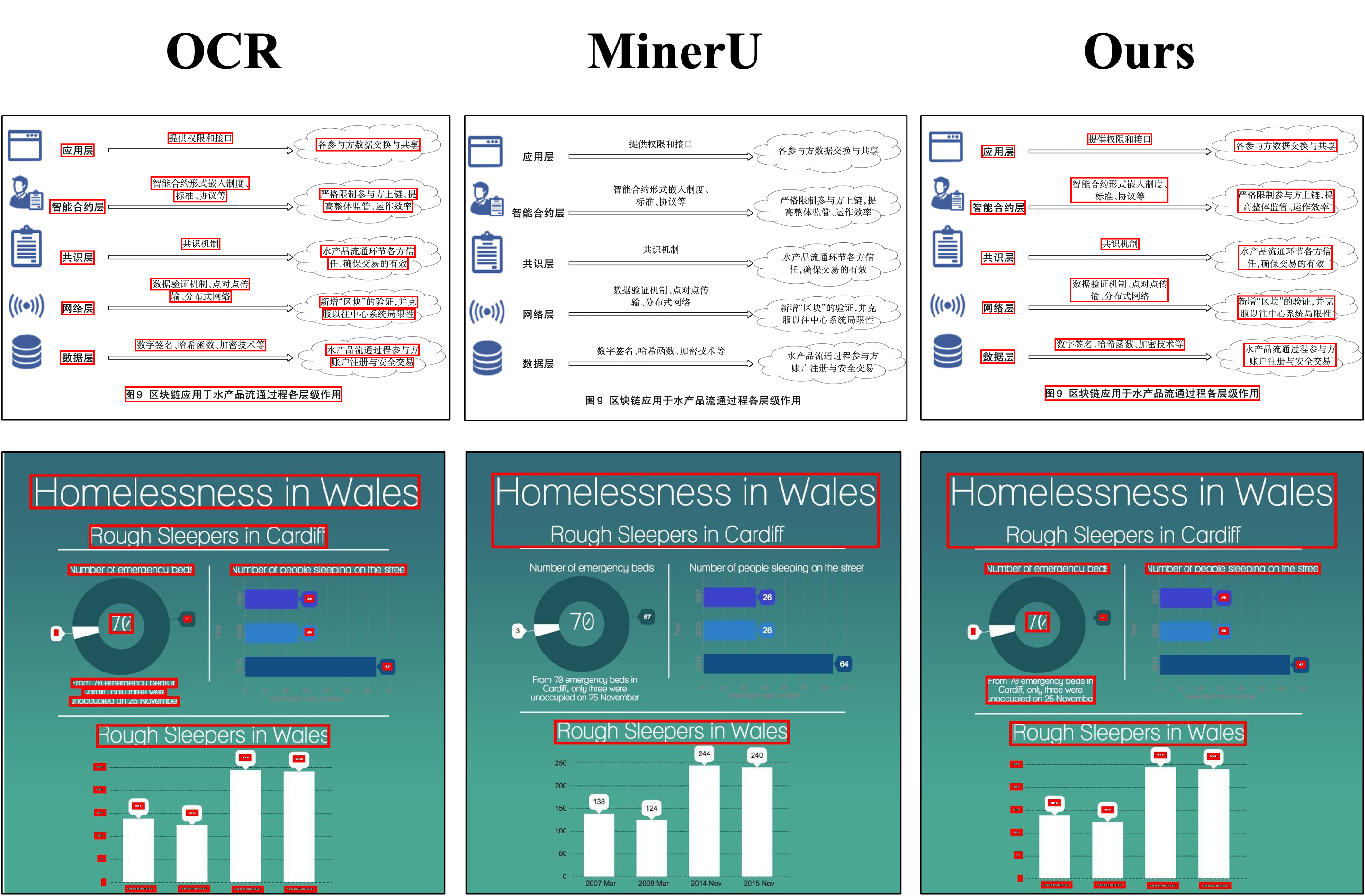} 
    \caption{Comparison of utilizing different strategies. EasyOCR offer line-by-line results ignoring semantic coherence, while MinerU often fail to identify some text-containing areas. Our proposed adaptive processing with refinement strategy can accurately handle images from different scenarios.}
    \label{img/process_compare}
    \vspace{0.4cm}
\end{figure}

We classify document and infographic as the hard scenarios, and the others as the easy scenarios. 
For images belonging to easy scenarios, we directly merge the OCR results based on their spatial relevance, the algorithm is shown in Table \ref{figures:Algorithm}.
As for images belonging to hard scenarios, we first employ MinerU \cite{minerU} to process the original images. MinerU is a specialized cascaded system designed for document-type pdf, which organizes the recognized content into blocks attached by corresponding bounding box and block type labels such as text, image or table. Nevertheless, it sometimes overlooks some text regions or misclassifies text-containing blocks as images.
To mitigate this issue, we refine MinerU’s output by leveraging the initial OCR results. Specifically, we extract a subset of initial OCR results that are omitted by MinerU through analysis of bounding box overlaps, and merge them based on spatial relevance. These recovered OCR results are then extended into MinerU’s output to supplement its original results.
As shown in Figure \ref{img/process_compare}, utilizing adaptive processing with refinement strategy can accurately handle images from different scenarios.

\begin{table}[]
\fontsize{9.0}{8.8}\selectfont
\centering
\renewcommand\arraystretch{1.3}
\begin{tabular*}{0.43\textwidth}{@{}l@{}}
\toprule
\textbf{Algorithm 1} Spatial merge
\\
\midrule

\textbf{Input}: OCR results $\mathcal{O}$, thresholds $x_{\text{ths}}$, $y_{\text{ths}}$ \\
\textbf{Output}: Merged text boxes $\mathcal{R}$ 

1: $\mathcal{B} \leftarrow \varnothing$ \\
2: \textbf{for} $b \in \mathcal{O}$: \\
3: \quad Store $(b_{\text{text}}, x_{\min}, x_{\max}, y_{\min}, y_{\max}, h, \bar{y}_c, 0)$ \\
4: \quad where $h = y_{\max} - y_{\min},\ \bar{y}_c = \frac{y_{\min} + y_{\max}}{2}$ \\

5: $g \leftarrow 1$ \\
6: \textbf{while} ungrouped boxes exist: \\
7: \quad \textbf{if} group $g$ empty: \\
8: \quad\quad Assign first ungrouped box to $g$ \\
9: \quad \textbf{else}: \\
10: \quad\quad Compute group bounds using $\pm x_{\text{ths}}\bar{h}, \pm y_{\text{ths}}\bar{h}$ \\
11: \quad\quad \textbf{for} ungrouped box $u$: \\
12: \quad\quad\quad \textbf{if} $u$ overlaps bounds: \\
13: \quad\quad\quad\quad Assign $u$ to $g$; \textbf{break} \\
14: \quad\quad \textbf{if} no assignment: $g \leftarrow g + 1$ \\

15: $\mathcal{R} \leftarrow \varnothing$ \\
16: \textbf{for} each group $k$: \\
17: \quad text $\leftarrow$ "" \\
18: \quad box $\leftarrow [0,0,0,0]$ \\
19: \quad \textbf{while} group boxes remain: \\
20: \quad\quad Find highest candidate row $\mathcal{S}$ \\
21: \quad\quad Select leftmost box $b^*$ in $\mathcal{S}$ \\
22: \quad\quad Append $b^*_{\text{text}}$ to text \\
23: \quad\quad Combine box $b^*_{\text{box}}$ and box \\
24: \quad Store merged text and group bbox \\
25: \quad $\mathcal{R} \leftarrow \mathcal{R} \cup \{(\text{text}: \text{text.strip}(), \text{box}: \text{box})\}$ \\
26: \textbf{return} $\mathcal{R}$ \\
\bottomrule
\end{tabular*}
\caption{Algorithm of spatial merge, which merges the OCR results based on their spatial relevance.}
\label{figures:Algorithm}
\end{table}

\subsubsection{Instruction Tuning Dataset}
Based on the processed results, we leverage GPT-4o \cite{DBLP:journals/corr/abs-2410-21276} to generate translations. Specifically, we prompt GPT-4o to generate 100 diverse questions for region-specific translation and full-image translation with grounding tasks, which are randomly sampled as questions for each instance. For each image, we construct a region-specific translation question-answer pair for each bounding box, and one full-image translation with grounding question-answer pair. The details of the templates are shown in Appendix \ref{sec:appendix_instruction_data}.

\section{PATIMT Benchmark}
In this section, we present a detailed description of our PATIMT-Bench. Firstly, we formally define the two sub-tasks mentioned above. Secondly, we conduct a comprehensive analysis of our datasets. The specific details of these two aspects are presented as follows.
\subsection{Task Definition}
 Our PATIMT-Bench focus on two sub-tasks:
\vspace{-0.2\topsep}
\begin{itemize}
    \setlength{\itemsep}{0.1cm}
    \setlength{\parsep}{1pt}
    \setlength{\parskip}{1pt}
    \item \textbf{Region-specific translation:}  Given an input image with specified bounding box coordinates in the prompt, the model needs to generate its accurate translation. 
    \item \textbf{Full-image translation with grounding:} Given an input image, the model needs to generate the text translation and the corresponding bounding box for each layout. This supports spatial correspondence between target text and source text and within the input image in practical applications. 
\end{itemize}
\vspace{-\partopsep}


\begin{table}[]
\centering
\begin{scriptsize}
\setlength{\tabcolsep}{1.5mm}
\renewcommand\arraystretch{1.5}
\begin{tabular*}{0.48\textwidth}{@{}l|ccccc@{}}
\toprule[1pt]

\textbf{PATIMT} & \textbf{Images} & \textbf{OCR boxes} & \textbf{Boxes} & \textbf{Src Words} & \textbf{Tgt Words} \\ \hline
\multicolumn{1}{l|}{Train} & 48,884
 & 1,307,516 & 417,066 & 24,827,252 & 30,437,907 \\
\multicolumn{1}{l|}{Test} & 1,200 &	- & 11,102 & 564,656 & 685,375
 \\

\bottomrule[1pt]
\end{tabular*}
\caption{Data statistics of PATIMT-Bench. OCR boxes, boxes refer to raw OCR box count and box count utilizing our pipeline, src words and tgt words refer to total number of words in source text and target text. Typically, test set is manually labeled.}
\label{figures:data_statistics}
\end{scriptsize}
\end{table}



\begin{table*}[t]
\fontsize{7.9}{8.8}\selectfont
\centering
\renewcommand\arraystretch{1.5}
\begin{tabular*}{0.965\textwidth}{@{}lcccccccccc}

\toprule[1pt]

\multicolumn{1}{l|}{}                                        & \multicolumn{4}{c|}{\textbf{Region-Specific Translation}}                                                          & \multicolumn{6}{c}{\textbf{Full-Image Translation with Grounding}}                                                   \\ 
\multicolumn{1}{l|}{}                                        & \multicolumn{2}{c}{\textbf{EN-ZH}} & \multicolumn{2}{c|}{\textbf{ZH-EN}}                                           & \multicolumn{3}{c}{\textbf{EN-ZH}}                & \multicolumn{3}{c}{\textbf{ZH-EN}}                \\
\multicolumn{1}{l|}{\multirow{-3}{*}{\textbf{Model}}}        & \textbf{BLEU}    & \textbf{COMET}  & \textbf{BLEU}   & \multicolumn{1}{c|}{\textbf{COMET}}                         & \textbf{BLEU}   & \textbf{COMET} & \textbf{IoU}   & \textbf{BLEU}   & \textbf{COMET} & \textbf{IoU}   \\ \hline
\multicolumn{11}{c}{\textit{\textbf{Proprietary LVLMs}}}        \\ \hline
\multicolumn{1}{l|}{Qwen2.5-VL-72B}                          & \underline{45.0}  & 76.6  & \textbf{37.3} & \multicolumn{1}{c|}{75.2}            & 13.2 & 48.7  & 0.185          & \underline{11.3}  & 53.0  & 0.251          \\
\multicolumn{1}{l|}{GPT-4o}   & 22.8 & 60.6 & 16.3 &  \multicolumn{1}{c|}{58.7} & 6.9 & 47.4 & 0.068 & 8.2 & 48.5 & 0.094 \\ \hline
\multicolumn{11}{c}{\textit{\textbf{Compact LVLMs}}}   \\ \hline
\multicolumn{1}{l|}{Aquila-VL-2B}    & 3.1  & 45.9  & 2.2    & \multicolumn{1}{c|}{44.0}       & 1.0  & 15.7          & 0.037          & 0.3 & 20.8          & 0.056          \\
\rowcolor[HTML]{EFEFEF} 
\multicolumn{1}{l|}{\cellcolor[HTML]{EFEFEF}Aquila-VL-2B*}    & 40.3           & 79.7           & 17.3          & \multicolumn{1}{c|}{\cellcolor[HTML]{EFEFEF}63.5}          & 19.6          & \underline{65.0}          & 0.359          & 7.4           & 53.1          & 0.332          \\
\multicolumn{1}{l|}{InternVL2.5-2B}       & 17.6          & 59.8           & 12.3          & \multicolumn{1}{c|}{58.1}                                  & 6.6           & 47.5          & 0.057          & 5.3           & 47.2          & 0.047          \\
\rowcolor[HTML]{EFEFEF} 
\multicolumn{1}{l|}{\cellcolor[HTML]{EFEFEF}InternVL2.5-2B*}  & 44.0           & 79.8           & 29.2          & \multicolumn{1}{c|}{\cellcolor[HTML]{EFEFEF}74.0}          & \underline{20.0}          & 59.6          & \underline{0.411}          & 10.9          & 54.6          & \underline{0.426}          \\
\multicolumn{1}{l|}{DeepseekVL2-Tiny}                          & 3.2           & 46.1           & 3.1           & \multicolumn{1}{c|}{50.9}                                  & 0.0          & 0.0          & 0.000          & 0.0           & 0.0          & 0.000          \\
\rowcolor[HTML]{EFEFEF} 
\multicolumn{1}{l|}{\cellcolor[HTML]{EFEFEF}DeepseekVL2-Tiny*} & \underline{43.8}           & \underline{85.4}           & \underline{29.8}          & \multicolumn{1}{c|}{\cellcolor[HTML]{EFEFEF}\underline{77.8}}          & 16.3          & 61.0          & 0.199          & \underline{11.2}          & \underline{57.3}          & 0.313          \\

\multicolumn{1}{l|}{SmoVLM2-2.2B}    & 2.1  & 41.2 & 1.1 & \multicolumn{1}{c|}{37.3} & 0.0 & 0.0 & 0.000  & 0.0 & 0.0 & 0.000 \\

\rowcolor[HTML]{EFEFEF} 
\multicolumn{1}{l|}{\cellcolor[HTML]{EFEFEF}SmoVLM2-2.2B*}  & 22.0 &  
66.4 & 11.6  & 
 \multicolumn{1}{c|}{\cellcolor[HTML]{EFEFEF}53.1} & 12.3 & 57.5 & 0.257 & 10.2 
& 51.2 & 0.235 \\

\multicolumn{1}{l|}{PaliGemma2-3B}    & 0.1 & 34.8 & 1.1 & \multicolumn{1}{c|}{37.6}  & 0.0 & 0.0 & 0.000 & 0.0 & 0.0 & 0.000   \\

\rowcolor[HTML]{EFEFEF} 
\multicolumn{1}{l|}{\cellcolor[HTML]{EFEFEF}PaliGemma2-3B*}    & 13.4 
& 55.4 & 24.8  & \multicolumn{1}{c|}{\cellcolor[HTML]{EFEFEF}65.4}          & 14.7 & 54.6 & 0.106 & 10.9 & 51.7 & 0.157  \\

\multicolumn{1}{l|}{Qwen2.5-VL-3B}                           & 19.5           & 63.0           & 10.5          & \multicolumn{1}{c|}{58.3}                                  & 3.3           & 19.6          & 0.073          & 2.2           & 17.8          & 0.068          \\
\rowcolor[HTML]{EFEFEF} 
\multicolumn{1}{l|}{\cellcolor[HTML]{EFEFEF}Qwen2.5-VL-3B*}  & \textbf{53.6}  & \textbf{87.7}  & \textbf{36.8}          & \multicolumn{1}{c|}{\cellcolor[HTML]{EFEFEF}\textbf{80.5}} & \textbf{26.4} & \textbf{67.0} & \textbf{0.457} & \textbf{17.5} & \textbf{59.4} & \textbf{0.427} \\ \hline
\multicolumn{11}{c}{\textit{\textbf{Cascade Pipelines}}}               \\ \hline
\multicolumn{1}{l|}{EasyOCR + LLM}                          & 21.3           & 58.0           & 19.1          & \multicolumn{1}{c|}{63.0}                                  & 5.5           & 47.0          & 0.223          & 5.4           & 49.0          & 0.305          \\ \multicolumn{1}{l|}{GOT-OCR + LLM}                                  & 38.2 & 75.5 & 27.3 & \multicolumn{1}{c|}{71.7} & 11.6 & 45.9 & - & 6.3 & 47.4 & - 
 \\ 
\bottomrule[1pt]

\end{tabular*}
\caption{Evaluation results of proprietary, compact LVLMs and cascade systems on PATIMT-Bench across two sub-tasks: region-specific translation and full-image translation with grounding, evaluated on both English $\xrightarrow{}$ Chinese (EN-ZH) and Chinese $\xrightarrow{}$ English (ZH-EN) using BLEU, COMET, and IoU metrics. Models marked with * indicate fine-tuning on our PATIMT train set. Best results are marked in \textbf{bold} and second-best results are \underline{underlined}. }
\label{MainExp_AVG}
\end{table*}

\subsection{Analysis of PATIMT-Bench }

\subsubsection{Training Dataset}

\begin{figure}[t]
    \centering \includegraphics[width=0.49\textwidth]
	{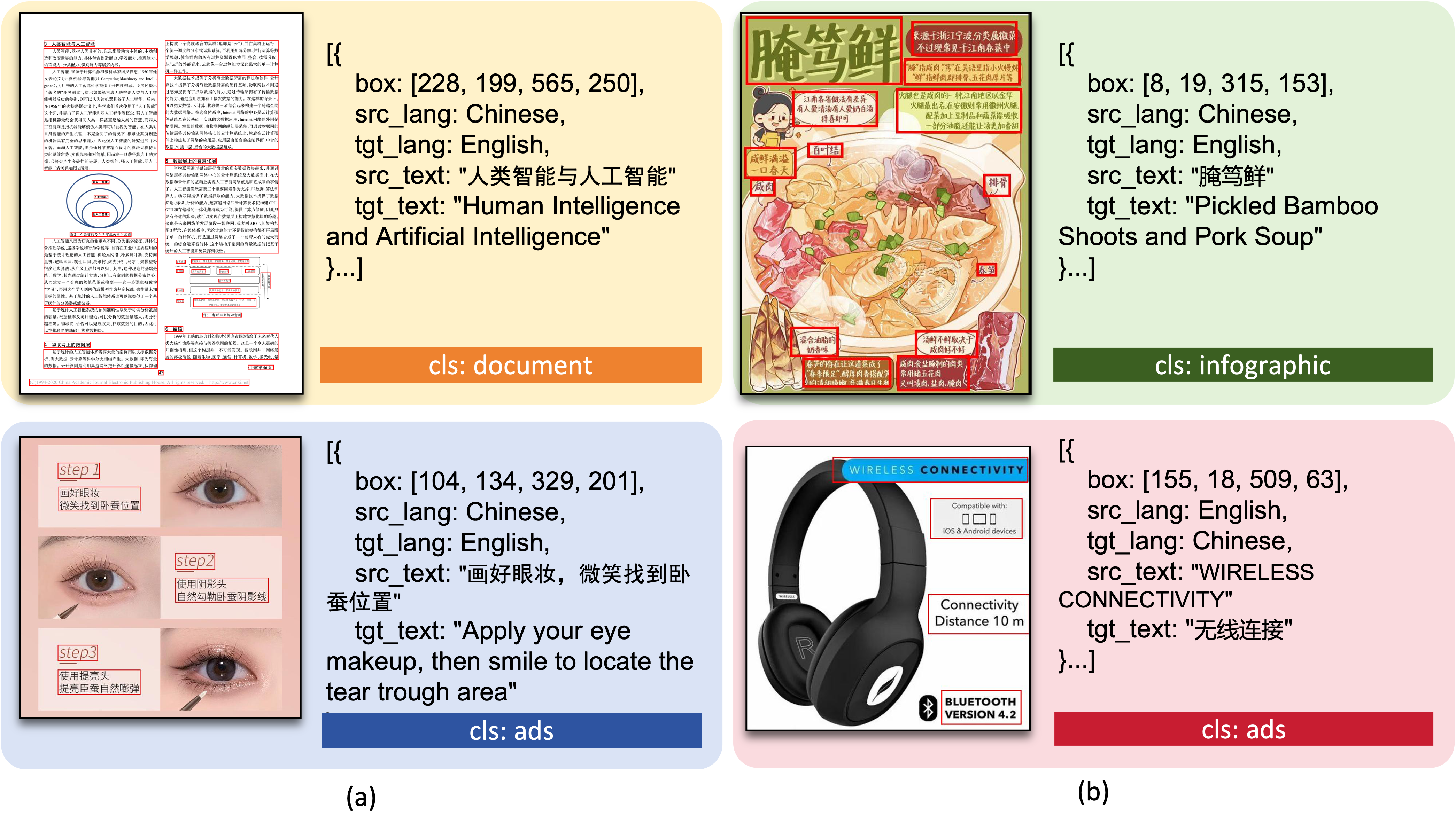} 
	\caption{samples from train set (left) and test set (right). }
	\label{img/datasample}
    \vspace{-0.1cm}
\end{figure}

We quantify key dataset metrics including image numbers, original OCR detection box numbers, box numbers after processing by our pipeline, and source/target word numbers, as presented in Table \ref{figures:data_statistics}.
A significant drop can be observed in the number of bounding boxes after processing, indicating that our pipeline effectively merges line-level OCR detection boxes to mitigate potential semantic fragmentation.
To ensure the quality of our dataset, we further conduct manual validation on a randomly sampled subset of 1,000 training images and verify both bounding box annotations and translations, which achieves a 92\% approval rate. We also exhibit some examples as shown in Figure \ref{img/datasample}.

\subsubsection{Test set}
Specifically, considering stylistic similarities in text across certain image categories, during the evaluation, we group advertisement/poster/book cover, table/chart, and natural scene/street view into three categories and remain the other scenarios unchanged. Therefore, the final evaluation includes images of six categories. The test set consists of 1,200 images including English $\xrightarrow{}$ Chinese and Chinese $\xrightarrow{}$ English, with 100 images manually selected and annotated for each category. Similar to the training set, the statistical results of the indicators and the sample demonstrations are shown in Table \ref{figures:data_statistics} and Figure \ref{img/datasample}.


\section{Experiments}

Section \ref{sec:setting} outlines our experimental setup, including evaluation metrics, baseline models and implementation details. Section \ref{sec:main_result} presents the main results, demonstrating performance improvements obtained by training on our dataset. In Section \ref{sec:ablation}, we conduct an ablation study to evaluate the effectiveness of our data construction pipeline. Section \ref{sec:scale} assesses the scalability of the training dataset, while Section \ref{sec:fox} examines its generalizability on the relevant benchmark. Finally, Section \ref{sec:pixel_time} provides an analysis of the trade‑off between speed and performance across varying image‑compression ratios.

\subsection{Experimental Setting}
\label{sec:setting}
\subsubsection{Metrics}

We report case-sensitive detokenized BLEU using SacreBLEU \cite{DBLP:conf/acl/PapineniRWZ02} and COMET \cite{DBLP:conf/emnlp/ReiSFL20} to evaluate translation quality, and assess the grounding capability in full-image translation with grounding task using the Intersection over Union (IoU) metric.

\subsubsection{Baselines} 

\textbf{Compact LVLMs.}\quad We select six LVLMs as our baselines: Aquila‑VL‑2B \cite{DBLP:journals/corr/abs-2410-18558}, InternVL‑2.5‑2B \cite{DBLP:journals/corr/internvl2.5}, Deepseek‑VL2‑Tiny \cite{DBLP:journals/corr/deepseekvl2}, SMOVLM2-2.2B\footnote{https://huggingface.co/blog/smolvlm2},
PaliGemma2-3B \cite{paligemma2} and Qwen2.5‑VL‑3B \cite{DBLP:journals/corr/qwen2.5-vl}. Each model is evaluated under two conditions: zero‑shot inference and fine‑tuning on our proposed training data. This dual evaluation provides a reliable assessment of the quality of our constructed dataset. Detailed introduction of these baseline models are listed in Appendix \ref{sec:appendix_baselinemodel}.

\textbf{Proprietary LVLMs.}\quad To benchmark our approach against more advanced vision–language models, we establish proprietary models using two state-of-the-art LVLMs: Qwen2.5‑VL‑72B \cite{DBLP:journals/corr/qwen2.5-vl} and GPT‑4o \cite{DBLP:journals/corr/abs-2410-21276}, both evaluated without fine‑tuning on our dataset.

\textbf{Cascade Pipelines.}\quad
Additionally, we implement cascade baselines that integrate EasyOCR or GOT-OCR \cite{GOT-OCR} for text recognition with Qwen2.5-VL-3B \cite{DBLP:journals/corr/qwen2.5} for translation to facilitate comparison with the aforementioned end-to-end image machine translation methods.

\begin{figure*}[t]
    \centering \includegraphics[width=0.98\textwidth]{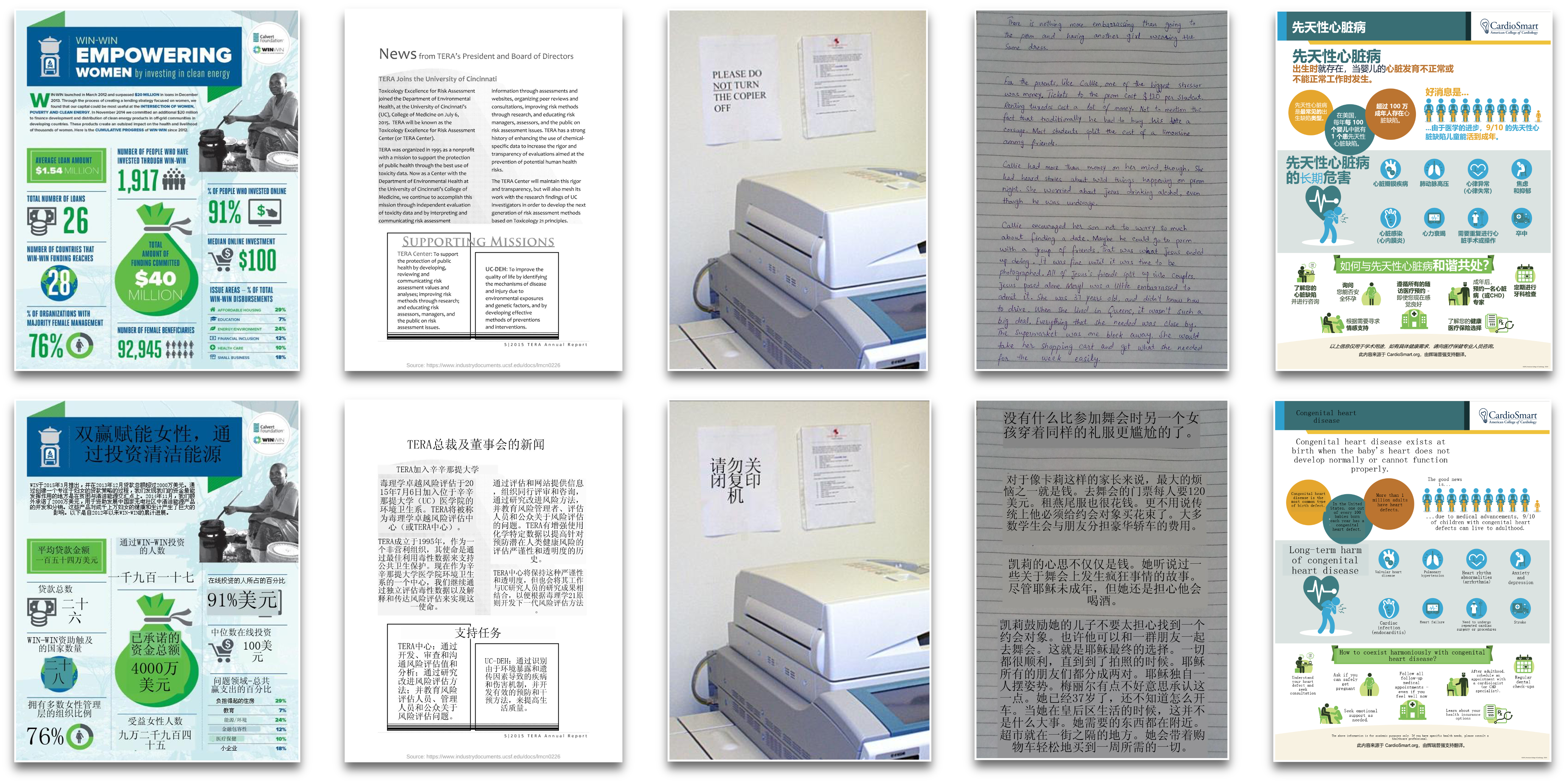} 
    \caption {Visualization of the full-image translation with grounding results by rendering the model outputs onto the corresponding source images based on their grounding information. The top row shows the source images, while the bottom row displays the rendered outputs.}
    \label{img/pred_showcase}
\end{figure*} 

\begin{table}
\begin{scriptsize}
\centering
\begin{subtable}{0.48\textwidth}
\centering
\setlength{\tabcolsep}{2.9mm}
\renewcommand\arraystretch{1.5}
\begin{tabular}{l|cccc}
\toprule[1pt]
\multirow{2}{*}{\textbf{Data Process}} & \multicolumn{2}{c}{\textbf{EN-ZH}} & \multicolumn{2}{c}{\textbf{ZH-EN}} \\
                               & \textbf{BLEU}   & \textbf{COMET}   & \textbf{BLEU}   & \textbf{COMET}   \\ \hline
OCR only                           & 42.0          & 79.4            & 29.8          & 75.6            \\
MinerU only   & 46.7 & 81.4 & 31.7 & 74.2 
            \\
Ours                              & \textbf{49.3}   & \textbf{84.0}    & \textbf{34.0}  & \textbf{76.7}       \\                 
\bottomrule[1pt]
\end{tabular}
\caption{Ablation on region-specific translation.}
\vspace{0.3cm}
\end{subtable}
\\

\begin{subtable}{0.48\textwidth}
\centering
\setlength{\tabcolsep}{1.35mm}
\renewcommand\arraystretch{1.5}
\begin{tabular}{l|cccccc}
\toprule[1pt]
\multirow{2}{*}{\textbf{Data process}} & \multicolumn{3}{c}{\textbf{EN-ZH}}         & \multicolumn{3}{c}{\textbf{ZH-EN}} \\
& \textbf{BLEU} & \textbf{COMET} & \textbf{IoU} & \textbf{BLEU} & \textbf{COMET} & \textbf{IoU} \\ \hline
OCR only & 16.0 & 58.8 & 0.319 
& 10.9 & 50.8 & 0.326  \\
MinerU only & 21.0 & 63.9 & 0.343 & 13.0 & 54.4 & 0.357
  \\
Ours  & \textbf{22.6} & \textbf{66.0} & \textbf{0.414} & \textbf{14.3} & \textbf{57.9} & \textbf{0.367}             
\\
\bottomrule[1pt]
\end{tabular}
\caption{Ablation on full-image translation with grounding.}
\end{subtable}
\caption{Ablation on different data processing methods. \textit{OCR only} and \textit{MinerU only} denote using EasyOCR and MinerU to generate OCR results without spatial merge and refinement. Ours denotes using our data construction pipeline. Best results are marked in \textbf{bold}.}
\label{figures:Ablation}
\end{scriptsize}
\end{table}

\subsubsection{Implementation Details}

Through our experiments, proprietary LVLMs, Qwen2.5‑VL-3B and InternVL‑2.5-2B can consistently generate outputs in JSON format, which yield superior performance, we fine‑tune Qwen2.5‑VL-3B and InternVL‑2.5-2B and evaluate all these four models in JSON format. The training of compact LVLMs is conducted with a batch size of 128 on four A6000 GPUs. Complete inference and training settings are provided in Appendix  \ref{sec:appendix_train_setting}.

\subsection{Main Results}
\label{sec:main_result}
Table \ref{MainExp_AVG} reports the average performance across different image scenarios in the PATIMT-Bench. Compact LVLMs struggle to follow PATIMT instructions under zero-shot settings, resulting in low BLEU, COMET, and IoU scores. After fine-tuning on our proposed dataset, all compact LVLMs achieve competitive performance, with some results even surpassing Qwen2.5-VL-72B and GPT-4o. For instance, the BLEU score of Aquila-VL-2B in the EN-ZH region-specific translation task increases from 3.1 to 40.3, and COMET improves from 45.9 to 79.7. Remarkably, Qwen2.5-VL-3B stands out among the baselines, outperforming both cascade pipelines and proprietary LVLMs by wide margins in most metrics. Detailed results for each scenario are provided in Appendix \ref{sec:appendix_exp_details}.

For clear visualization, we render the results of full-image translation with grounding based on the predicted grounding information, as shown in Figure \ref{img/pred_showcase}, which demonstrates that our model generates translations with accurate grounding.

Overall, these results validate the effectiveness of our dataset in enhancing translation quality and text spatial grounding to handle PATIMT.


\begin{table}[t]
\begin{scriptsize}

\begin{subtable}{.5\textwidth}
\centering
\setlength{\tabcolsep}{3.6mm}
\renewcommand\arraystretch{1.5}
\begin{tabular}{@{}l|cccc}
\toprule[1pt]
\multirow{2}{*}{\textbf{Scale}} & \multicolumn{2}{c}{\textbf{EN-ZH}} & \multicolumn{2}{c}{\textbf{ZH-EN}} \\
                                     & \textbf{BLEU}   & \textbf{COMET}   & \textbf{BLEU}   & \textbf{COMET}   \\ \hline
5K                                   & 49.3          & 84.0            & 34.0          & 76.7            \\
10K                                  & 51.0          & 84.3            & 35.2          & 77.9            \\
all                                  & \textbf{53.7}  & \textbf{87.7}   & \textbf{36.8}  & \textbf{80.5}     \\      
\bottomrule[1pt]
\end{tabular}
\caption{Results on region-specific translation.}
\end{subtable}
\\

\begin{subtable}{.5\textwidth}
\setlength{\tabcolsep}{1.8mm}
\renewcommand\arraystretch{1.5}
\centering

\begin{tabular}{@{}l|cccccc}
\toprule[1pt]
\multirow{2}{*}{\textbf{Scale}} & \multicolumn{3}{c}{\textbf{EN-ZH}}            & \multicolumn{3}{c}{\textbf{ZH-EN}}            \\
& \textbf{BLEU} & \textbf{COMET} & \textbf{IoU} &\textbf{BLEU} & \textbf{COMET} & \textbf{IoU} \\ \hline
5K & 22.6 & 63.0 & 0.414 & 14.3 & 57.9 & 0.367 
\\
10K & 24.3 & 64.7 & 0.432 & 15.8 & 58.7 & 0.405 
\\
all & \textbf{26.4} & \textbf{67.0} & \textbf{0.457} & \textbf{17.5} & \textbf{59.4} & \textbf{0.427} 
\\
\bottomrule[1pt]
\end{tabular} 
\caption{Results on full-image translation.}
\end{subtable}
\vspace{-0.4cm}
\end{scriptsize}
\caption{Results of scalability of our data based on Qwen2.5-VL-3B, $d$K denotes the base model is fine-tuned on $d$K subset from our train set, \textit{all} represents training on the entire dataset. The best results are marked in \textbf{bold}.}
\label{figures:Scale}
\end{table}

\subsection{Ablation Study}
\label{sec:ablation}

To evaluate the effectiveness of our data construction pipeline, we assess the performance of Qwen2.5-VL-3B using EasyOCR or MinerU annotations without implementing our adaptive processing with refinement strategy.
Given the high cost of GPT-based labeling, we randomly sample 10 \% of instances from each scenario in our training dataset, resulting in a subset of 5,000 examples.
As shown in Table \ref{figures:Ablation}, the first row and second row represent training on the subset annotated by EasyOCR and MinerU without spatial merge and refinement. The last row corresponds to training on the subset processed by our pipeline, demonstrating a clear performance improvement.

\subsection{Scalability}
\label{sec:scale}

To assess the scalability of our dataset, we construct two additional training subsets of 5,000 and 10,000 instances. As illustrated in Table \ref{figures:Scale}, we observe a steady improvement in performance, thereby demonstrating the scalability of our dataset. 

\subsection{Extending to Other Benchmarks}
\label{sec:fox}

\begin{table}[]
\centering
\begin{small}
\setlength{\tabcolsep}{3mm}
\renewcommand\arraystretch{1.4}
\begin{tabular*}{0.38\textwidth}{@{}l|cccc@{}}
\toprule[1pt]

\textbf{Models}                     & \textbf{BLEU} & \textbf{COMET} \\ \hline
\multicolumn{1}{l|}{Fox}            & 13.8          & 36.6          \\
\multicolumn{1}{l|}{Qwen2.5-VL-3B}  & 9.4           & 89.2          \\
\multicolumn{1}{l|}{Qwen2.5-VL-3B*} & \textbf{47.9}          & \textbf{91.7}         
\\
\bottomrule[1pt]
\end{tabular*}
\caption{Comparison of our fine-tuned model on Fox benchmark \cite{DBLP:journals/corr/fox}. The best results are marked in \textbf{bold}. }
\label{figures:foxbench}
\end{small}
\vspace{-0.2cm}
\end{table}

To further assess the generalizability of our training data, we evaluate our baseline models on the Fox benchmark \cite{DBLP:journals/corr/fox}, which contains document region-specific text image machine translation. As shown in Table \ref{figures:foxbench}, our fine‑tuned model substantially outperform both the Fox model and the baseline model by approximately 400\% in BLEU score and 250\% in COMET, demonstrating the broad applicability of our training data and benchmark.

\subsection{Tradeoff Between Speed and Performance Across Image Compression Ratios.}
\label{sec:pixel_time}

\begin{figure}[t]
    \centering 
    \includegraphics[width=0.5\textwidth]{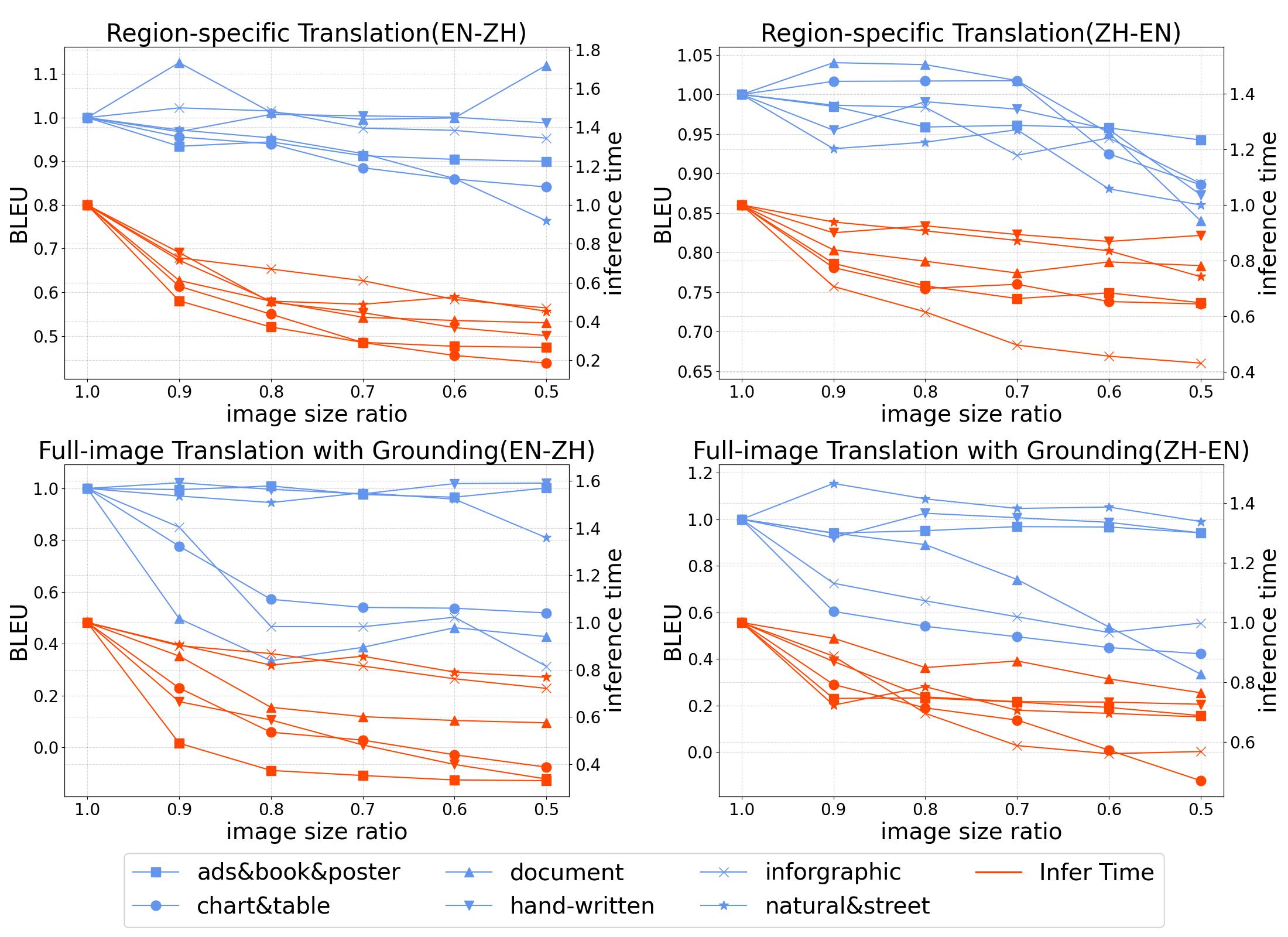} 
    \caption {Illustration of the BLEU score (left y-axis) and inference time (right y-axis) across varying compression ratios (x-axis).}
    \label{img/pixel_time}
\end{figure}
 High-resolution images provide fine-grained visual information facilitating great performance, while generate excessive visual tokens which significantly increase the inference time \cite{monkey}.
To explore this influence in PATIMT task, we conduct a series of experiments.
Specifically, we here compress the images in our test set to different ratios and measure the change of both inference time and BLEU score, as shown in Figure \ref{img/pixel_time}. For clarity, We report BLEU scores and inference times as decimal fractions of the uncompressed baseline performance. From the resulting plot, we draw the following conclusions:
\vspace{-0.3\topsep}
\begin{itemize}
    \setlength{\itemsep}{0.1cm}
    \setlength{\parsep}{1pt}
    \setlength{\parskip}{1pt}
    \item In region-specific translation task, the performance remains comparable even when images are significantly compressed. It indicates that this task holds the great potential to accelerate through the compression of visual features.
    \item In full-image translation with grounding task, the performance of some scenarios such as chart\&table, document, and infographic are sensitive to image compression ratio. These scenarios may be adversely affected by limited image size, whereas other categories maintain stable performance.
\end{itemize}

\section{Conclusion}
In this paper, we extend the conventional TIMT task into PATIMT task, which encompasses two sub-tasks: region-specific translation and full-image translation with grounding. Confronted with data scarcity, we construct the PATIMT-Bench, a benchmark featuring 10 distinct image scenarios. We introduce an Adaptive Image OCR Refinement Pipeline to construct training data, which adaptively selects suitable OCR tools according to different image scenarios and refines the results for text-rich images to ensure high-quality annotations. Notably, to ensure the accuracy of evaluation, we manually annotate bounding boxes and review the translation results of 1,200 instances to construct the test set. LVLMs fine-tuned on our data achieve state-of-the-art performance on PATIMT-Bench, and demonstrate the scaling ability of our training data.
In the future, we will explore the application of our dataset to domains including in-image machine translation \cite{tian-etal-2023-image,lan-etal-2024-translatotron} and visual text generation \cite{tuo2023anytext,liu2023culturalgaptexttoimagegeneration,li-etal-2024-empowering-backbone,sd3}. Additionally, we aim to expand our benchmark into a large-scale multilingual version.

\section*{Limitations}
Despite the contributions of our benchmark in advancing PATIMT and achieving impressive performance, several limitations still remain.  Our benchmark predominantly focuses on bounding boxes for region annotation. However, in practical applications, users may prefer or require other formats such as polygons, points, or free-form shapes. Besides, multilingual translation is not explored in our benchmark.

\section*{Acknowledgments}

The project is supported by 
National Key R\&D Program of China (No. 2022ZD0160501), 
Natural Science Foundation of Fujian Province of China (No. 2024J011001),
and
the Public Technology Service Platform Project of Xiamen (No.3502Z20231043).
We also thank the reviewers for their insightful comments.

\bibliography{acl_latex}

\clearpage

\appendix

\section{Adaptive Image OCR Refinement
Pipeline}
\label{sec:appendix_pipeline}

\subsection{Data Comparison}
\label{sec:appendix_data_compare}
For more clear comparison, we show the example input images and the output format of our dataset comparing to existing TIMT datasets, as shown in Figure \ref{figures:data_compare_vis}.

\subsection{CLIP-based Categorization}
\label{sec:appendix_clip_cls}
Following \cite{DBLP:journals/corr/abs-2306-17107}, we divide the images in our training data into 10 distinct classes. Each class is associated with one or more descriptive labels.
\begin{itemize}
\setlength{\itemsep}{0.1cm}
\setlength{\parsep}{1pt}
\setlength{\parskip}{1pt}
    \item \textbf{ads}: "advertisement"
    \item \textbf{book}: "book cover", "magazine cover", "comic book cover"
    \item \textbf{poster}: "movie poster", "podcast poster", "TV show poster", "event poster", "poster", "concert poster", "conference poster", "travel poster", "art poster"
    \item \textbf{natural}: "natural scene", "landscape", "nature background", "wildlife scene", "Trail sign", "Park map", "Info board", "Gate sign", "Stone plaque", "Wood post","Kiosk sign", "Exhibit panel"
    \item \textbf{street}: "street view", "urban scene", "city street", "suburban neighborhood", "rural road", "traffic scene", "billboard", "shop front"
    \item \textbf{hand-written}: "hand-written", "handwriting letter"
    \item \textbf{infographic}: "infographic", "diagram", "mind map", "statistical graph"
    \item \textbf{document}: "document", "contract"
    \item \textbf{chart}: "chart", "bar chart", "pie chart", "scatter plot", "line chart", "Histogram", "area chart", "bubble chart",
    \item \textbf{table}: "table", "spreadsheet", "matrix", "grid"
\end{itemize}

For each word, we apply the same textual templates used in \citet{DBLP:journals/corr/abs-2306-17107} to achieve embedding-space ensembling \cite{DBLP:conf/icml/clip}:
\begin{itemize}
\setlength{\itemsep}{0.1cm}
\setlength{\parsep}{1pt}
\setlength{\parskip}{1pt}
    \item "a photo of a \{\}.",
    \item "a blurry photo of a \{\}.",
    \item "a black and white photo of a \{\}.",
    \item "a low contrast photo of a \{\}.",
    \item "a high contrast photo of a \{\}.",
    \item "a bad photo of a \{\}.",
    \item "a good photo of a \{\}.",
    \item "a photo of a small \{\}.",
    \item "a photo of a big \{\}."
\end{itemize}

Using CLIP-ViT-L/14, we compute the similarity between each image and all associated labels. Each image is then assigned to the corresponding superclass (e.g., book) of the label (e.g., "book cover") with the highest similarity score.

\begin{table}[t]
\centering
\begin{scriptsize}
\setlength{\tabcolsep}{1.3mm}
\renewcommand\arraystretch{1.5}
\begin{tabular*}
{0.42\textwidth}
{lm{3cm}<{\centering} m{1.5cm}<{\centering}}
\toprule[1pt]
\textbf{Dataset} & \textbf{Input Image} & \textbf{Output Format} \\ \hline
\\
\multirow{1}*{OCRMT30K 
} &  {\centering\includegraphics[width=0.6\linewidth]{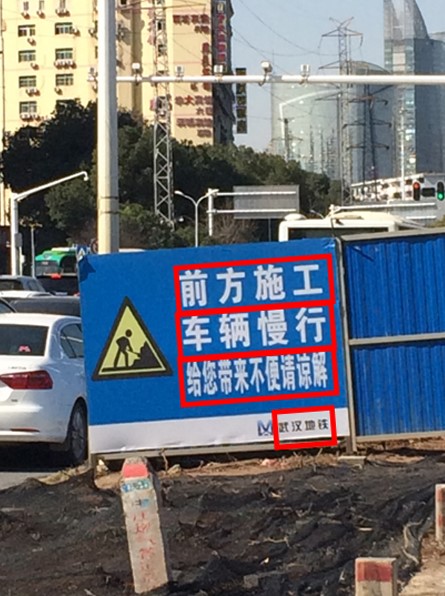}} & Plain Text \\ 

\multirow{1}*{DiTrans 
} &  {\centering\includegraphics[width=0.6\linewidth]{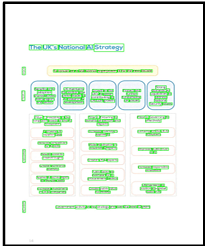}} & Plain Text \\ 

\multirow{1}*{DoTA 
} & {\centering\includegraphics[width=0.6\linewidth]{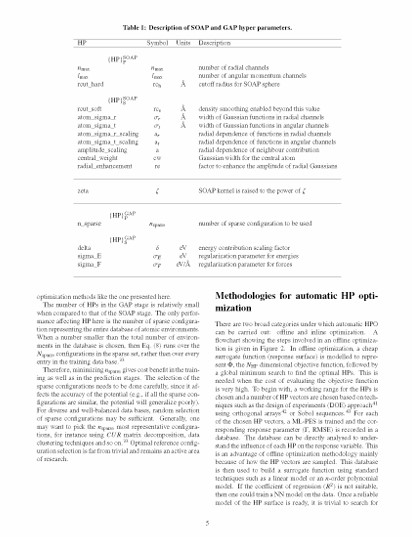}} & Markdown Text  \\ 
\multirow{1}*{UMTIT 
} & {\centering\includegraphics[width=0.6\linewidth]{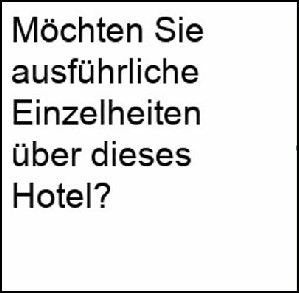}} & Image  
\\ 
\multirow{1}*{MIT-10M 
} & {\centering\includegraphics[width=0.6\linewidth]{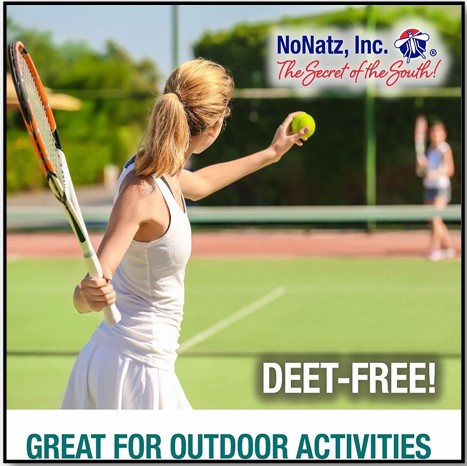}} &	Plain Text  \\ 
\multirow{1}*{PATIMT(Ours)} & {\centering\includegraphics[width=0.6\linewidth]{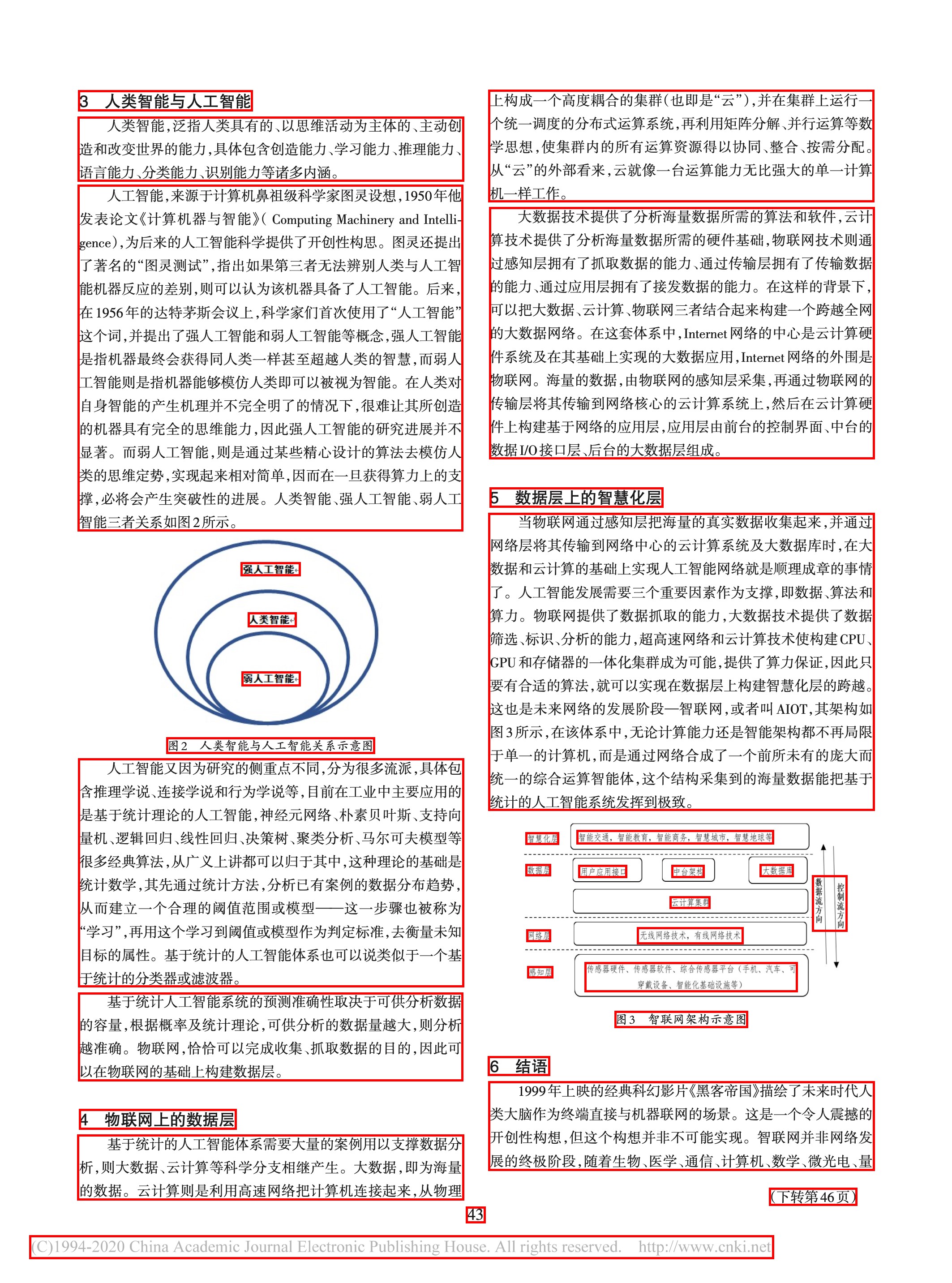}} & \makecell[c]{Plain Text \&\\ Bounding Box}  \\

\bottomrule[1pt]
\end{tabular*}
\caption{Image and output format comparison of PATIMT with other popular image translation datasets.}
\label{figures:data_compare_vis}
\end{scriptsize}
\end{table}

\subsection{Instruction Tuning Data}
\label{sec:appendix_instruction_data}
In this section, we detail the question and label formats used for each baseline model during training, as illustrated in Table \ref{figures:instruction_tuning_data}.


\begin{table*}[]
\centering
\begin{small}
\begin{sloppypar}
\begin{CJK}{UTF8}{gbsn}
\begin{subtable}{\textwidth}
\centering
\setlength{\tabcolsep}{1.8mm}
\renewcommand\arraystretch{1.7}
\begin{tabular}{m{2cm}<{\centering} m{7.5cm} m{5cm}}
\toprule[1pt]

\textbf{Image}                 & \textbf{Question Format} & \textbf{Response Format} \\ \hline
\multirow{2}*{\includegraphics[width=0.9\linewidth]{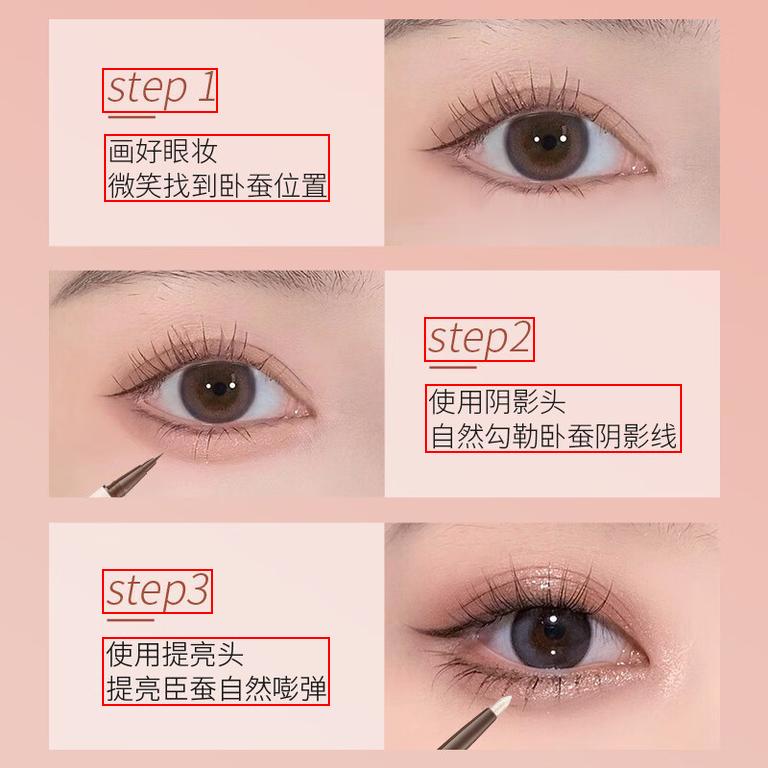}} & \textbf{First pinpoint the words in \textcolor{red}{Box([0.13, 0.09, 0.28, 0.15])}, then express them in Chinese.} Output only the recognized text content and translation result in format: text <|translation|> translation. & step1 <|translation|> 步骤1  \\
\cline{2-3}
~ & \textbf{Extract all visible text and offer its English meaning.} Return the recognized text content, translation result and boxes in format: text <|translation|> translation Box([x1, y1, x2, y2]). &  画好眼妆微笑找到卧蚕位置 <|translation|> Apply your eye makeup, then smile to locate the tear trough area Box([0.14, 0.17, 0.43, 0.26])$\backslash$n  ... \\
\bottomrule[1pt]
\end{tabular}
\caption{Example of instruction tuning data with plain-text format. Aquila-VL-2B and DeepseekVL2-Tiny utilize this format.}
\end{subtable}
\\

\begin{subtable}{0.95\textwidth}
\centering
\setlength{\tabcolsep}{1.8mm}
\renewcommand\arraystretch{1.5}
\begin{tabular}{m{2cm}<{\centering} m{8cm} m{4cm}}
\toprule[1pt]

\textbf{Image}                 & \textbf{Question Format} & \textbf{Response Format} \\ \hline
\multirow{2}*{\includegraphics[width=0.9\linewidth]{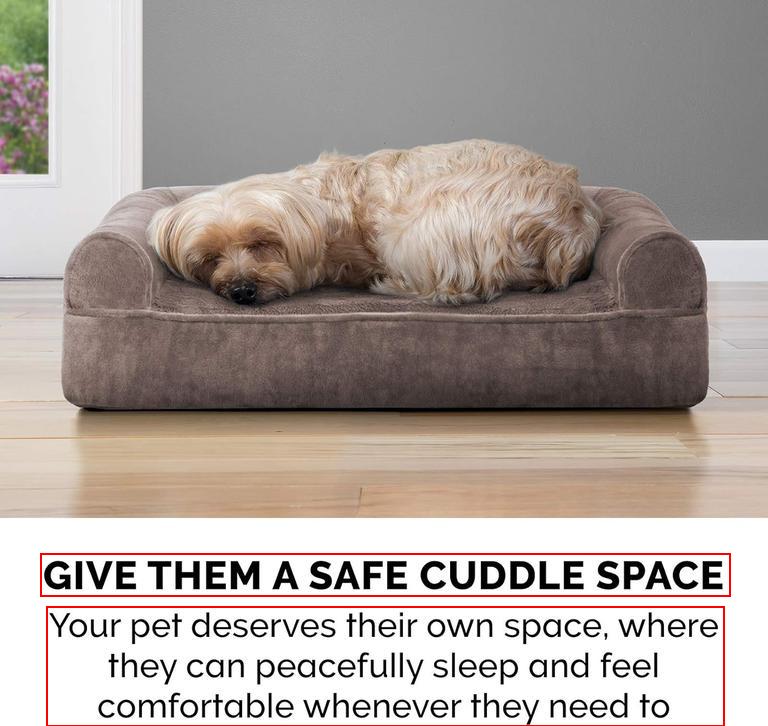}} & \textbf{First read the English snippet at \textcolor{red}{Box([40, 553, 730, 596])}, then provide its Chinese version.}. Output result in the following JSON format (note xxx is placeholder for text, x1,y1,x2,y2 are placeholders for coordinate).\{"bbox\_2d": Box([ x1,y1,x2,y2]), "text\_content": xxx, "translation": xxx\} & json\{"bbox\_2d": "Box([40, 553, 730, 596])","text\_content": "GIVE THEM A SAFE CUDDLE SPACE","translation": "给他们一个安全的拥抱空间"\} \\
\cline{2-3}
~ & \textbf{Can you do text detection and translation from English to Chinese?}. Output result in the following JSON format (note xxx is placeholder for text, x1,y1,x2,y2 are placeholders for coordinate, ... means there may be more contents in the image).[{"bbox\_2d": Box([x1,y1,x2,y2]), "text\_content": xxx, "translation": xxx},...]. &  json[\{"bbox\_2d": "Box([40, 553, 730, 596])","text\_content": "GIVE THEM A SAFE CUDDLE SPACE","translation": "给他们一个安全的拥抱空间"\}, ...] \\
\bottomrule[1pt]
\end{tabular}
\caption{Example of instruction tuning data with JSON format. InternVL2.5-2B and Qwen2.5VL-3B utilize this format.}
\end{subtable}

\end{CJK}
\end{sloppypar}
\end{small}
\caption{Example of instruction tuning data with different format. texts marked in \textbf{bold} refer to diverse question generated by GPT-4o, Box(·) denotes converting bounding box to the format utilized by each baseline model, such that Box([10,20,30,40]) is [10,20,30,40] for Qwen2.5-VL-3B and <box>[[10,20,30,40]]</box> for InternVL2.5-3-2B. }
\label{figures:instruction_tuning_data}
\vspace{-0.2cm}
\end{table*}

\section{Detailed Experiment Settings}
\subsection{Details of baseline models.}
\label{sec:appendix_baselinemodel}
We introduce the trainable parameters, bounding box format and other settings of our selected LVLM baseline models as the following:

\begin{itemize}
    \item \textit{Aquila‑VL‑2B} \cite{DBLP:journals/corr/abs-2410-18558}.\quad This model is developed based on the LLaVA-One-Vision framework \cite{DBLP:journals/corr/llava-ov}, utilizing the Qwen2.5-1.5B-Instruct \cite{DBLP:journals/corr/qwen2.5} as the language model and SigLIP-SO400M-Patch14-384\footnote{https://huggingface.co/google/siglip-so400m-patch14-384} as the vision tower. It contains a total of 2.18 billion trainable parameters. The bounding box format is [x1, y1, x2, y2], where each coordinate represents a normalized ratio in the range [0,1].

    \item \textit{InternVL‑2.5‑2B} \cite{DBLP:journals/corr/internvl2.5}.\quad This model employs InternLM2.5-1.8B-Chat \cite{cai2024internlm2technicalreport} as the large language model and InternViT-300M-448px-V2.5\footnote{https://huggingface.co/OpenGVLab/InternViT-300M-448px-V2\_5} as the vision tower, with a randomly initialized MLP projector. It has 2.21 billion trainable parameters. The bounding box format is <box>[x1, y1, x2, y2]</box>, where the coordinates are normalized to the range [0,1000].

    \item \textit{DeepSeek‑VL2‑Tiny} \cite{DBLP:journals/corr/deepseekvl2}.\quad This model is based on DeepSeekMoE-3B \cite{dai2024deepseekmoeultimateexpertspecialization}, comprising 3.37 billion trainable parameters and 1.0 billion activated parameters during inference. The bounding box format is <|det|>[x1, y1, x2, y2]<|/det|>, where coordinates are normalized to the range [0,999].

    \item \textit{SmoVLM2-2.2B}. \quad This model is designed for efficient video understanding across various devices, offering strong visual understanding and localization capabilities. The bounding box format is [x1, y1, x2, y2], where each coordinate represents a normalized ratio in the range [0,1].

    \item \textit{PaliGemma2-3B} \cite{paligemma2}. \quad This model connects the SigLIP image encoder with the Gemma2 language model, supporting various input resolutions (224x224, 448x448, and 896x896) for different use cases. The bounding box format is [x1, y1, x2, y2], where each coordinate represents a normalized ratio in the range [0,1].

    \item \textit{Qwen2.5‑VL‑3B} \cite{DBLP:journals/corr/qwen2.5-vl}.\quad This model demonstrates strong visual understanding and localization capabilities. It has 3.75 billion trainable parameters. The bounding box format is [x1, y1, x2, y2], using absolute position coordinates.
\end{itemize}

\subsection{Details of Training and Inference Configuration}
\label{sec:appendix_train_setting}
We list the detailed training settings as the following: 
\begin{itemize}
\item \textbf{Optimization Settings:}
    \begin{itemize}
        \item Learning rate: 
        1e-5 with cosine scheduling
        \item Warmup ratio: 0.1
        \item Weight decay: 0.0
        \item Batch size: 128
        \item Training epoch: 1
        \item Optimizer: AdamW
    \end{itemize}
    
    \item \textbf{Computational Environment:}
        \begin{itemize}
            \item Precision: bfloat16 (bf16)
            \item Acceleration framework: DeepSpeed Stage 3
            \item Hardware: 4× NVIDIA A6000 GPUs
        \end{itemize}
\end{itemize}

During inference, we set the temperature parameter to zero and employ greedy decoding. To prevent premature truncation of generated sequences, we specify the maximum number of new tokens as the greater of the ground-truth sequence or length 4096.

\section{Complete Results of Main Experiments}
\label{sec:appendix_exp_details}

\begin{table*}[]
\centering
\begin{scriptsize}

\begin{subtable}{0.98\textwidth}
\setlength{\tabcolsep}{1.55mm}
\renewcommand\arraystretch{1.3}
\begin{tabular}{lcccccccccccc@{}}
\toprule[1pt]
& \multicolumn{2}{c}{\textbf{ads\&book\&poster}} & \multicolumn{2}{c}{\textbf{chart\&table}} & \multicolumn{2}{c}{\textbf{document}} & \multicolumn{2}{c}{\textbf{hand-written}} & \multicolumn{2}{c}{\textbf{infographics}} & \multicolumn{2}{c}{\textbf{natural\&street}} \\
\multirow{-2}{*}{\textbf{Model}}           & \textbf{BLEU} & \textbf{COMET} & \textbf{BLEU} & \textbf{COMET} & \textbf{BLEU} & \textbf{COMET} & \textbf{BLEU} & \textbf{COMET} & \textbf{BLEU} & \textbf{COMET} & \textbf{BLEU} & \textbf{COMET}  \\ \hline
\multicolumn{13}{c}{\textit{\textbf{Proprietary LVLMs}}}      \\ \hline

Qwen2.5-VL-72B 
& 47.2 & 78.3 & 50.5 & 78.5 & 41.1 & 76.0 & 37.6 & 74.0 & 36.6 & 71.1 & 56.8 & 81.5 
   \\
GPT-4o  
&  36.6  & 72.8  & 16.2  & 54.5  & 8.1  & 45.3  & 28.8  & 68.3  & 11.7  & 51.6  & 35.4  & 70.9 
          \\ \hline
\multicolumn{13}{c}{\textit{\textbf{Compact LVLMs}}}  \\ \hline
Aquila-VL-2B 
& 2.7 & 50.1 & 2.7 & 43.1 & 2.1 & 40.5 & 3.4 & 41.7 & 2.4 & 44.1 & 5.3 & 55.6 
    \\
\rowcolor[HTML]{EFEFEF} Aquila-VL-2B* & 39.6 & 80.9 & 47.0 & 81.4 & 38.2 & 79.4 & 36.1 & 81.1 & 34.2 & 76.5 & 46.9 & 79.0 
   \\
InternVL2.5-2B                        & 23.3 & 68.4 & 13.3 & 54.3 & 8.6 & 48.7 & 20.2 & 64.8 & 10.6 & 53.3 & 29.7 & 69.0 
    \\
\rowcolor[HTML]{EFEFEF}InternVL2.5-2B*  & 49.8 & 81.8 & 47.5 & 83.2 & 39.4 & 79.3 & 36.8 & 76.5 & 38.7 & 77.8 & 52.0 & 80.0 
 \\
DeepseekVL2-Tiny   & 2.4 & 49.9 & 2.4 & 41.8 & 2.0 & 41.7 & 4.4 & 44.0 & 3.8 & 44.8 & 4.4 & 54.5 
  \\
\rowcolor[HTML]{EFEFEF}DeepseekVL2-Tiny* & 46.7 & 86.2 & 54.5 & 86.7 & 41.2 & 83.9 & 35.8 & 85.1 & 33.8 & \textbf{88.1} & 51.0 & 82.6           \\
SmolVLM2-2.2B & 3.9 & 47.9 & 1.2 & 38.2 & 1.2 & 34.6 & 0.5 & 37.8 & 1.6 & 37.5 & 4.2 & 51.2 \\

\rowcolor[HTML]{EFEFEF}SmolVLM2-2.2B* 
& 26.3 & 72.3 & 22.9 & 66.6 & 21.1 & 66.2 & 15.8 & 62.0 & 17.9 & 61.8 & 27.8 & 69.3 \\

PaliGemma2-3B & 0.3 & 43.5 &  0.1 & 34.6 & 0.1 & 28.3 & 0.0 & 25.0 & 0.1 & 34.6 & 0.0 & 42.8 \\

\rowcolor[HTML]{EFEFEF}PaliGemma2-3B* & 36.6 & 76.3 & 23.4 & 65.4 & 15.0 & 59.0 & 12.9 & 56.2 & 20.9 & 62.1 & 39.9 & 73.4 \\

Qwen2.5-VL-3B    & 31.3 & 68.2 & 20.9 & 62.2 & 8.6 & 56.0 & 6.0 & 61.9 & 12.2 & 53.7 & 38.0 & 75.7 
  \\
\rowcolor[HTML]{EFEFEF}Qwen2.5-VL-3B*  & \textbf{52.6} & \textbf{86.5} & \textbf{60.3} & \textbf{90.2} & \textbf{51.3} & \textbf{86.8} & \textbf{52.7} & \textbf{89.7} & \textbf{47.0} & 86.4 & \textbf{57.8} & \textbf{86.7} 
      \\ \hline
\multicolumn{13}{c}
{\textit{\textbf{Cascade Pipelines}}}                  \\ \hline EasyOCR + LLM & 20.2 & 59.0 & 23.6 & 63.3 & 31.6 & 69.3 & 3.8 & 32.8 & 34.2 & 69.2 & 14.5 & 54.6 \\ GOT-OCR + LLM & 40.9 & 77.4 & 41.3 & 76.2 & 33.1 & 73.1 & 43.6 & 80.1 &  32.8 & 73.3 &  37.5 & 73.1 

\\ \bottomrule[1pt]
\end{tabular}
\caption{Detailed results for region-specific translation task (EN-ZH).}
\vspace{2mm}
\end{subtable}
\\

\begin{subtable}{0.98\textwidth}
\setlength{\tabcolsep}{1.55mm}
\renewcommand\arraystretch{1.3}
\begin{tabular}{lcccccccccccc@{}}
\toprule[1pt]
& \multicolumn{2}{c}{\textbf{ads\&book\&poster}} & \multicolumn{2}{c}{\textbf{chart\&table}} & \multicolumn{2}{c}{\textbf{document}} & \multicolumn{2}{c}{\textbf{hand-written}} & \multicolumn{2}{c}{\textbf{infographics}} & \multicolumn{2}{c}{\textbf{natural\&street}} \\
\multirow{-2}{*}{\textbf{Model}}           & \textbf{BLEU} & \textbf{COMET} & \textbf{BLEU} & \textbf{COMET} & \textbf{BLEU} & \textbf{COMET} & \textbf{BLEU} & \textbf{COMET} & \textbf{BLEU} & \textbf{COMET} & \textbf{BLEU} & \textbf{COMET}  \\ \hline
\multicolumn{13}{c}{\textit{\textbf{Proprietary LVLMs}}}      \\ \hline

Qwen2.5-VL-72B 
& 30.4 & 73.5 & 48.7 & 83.6 & 40.3 & 76.2 & 28.0 & 67.3 & 46.6 & 81.4 & 29.6 & 69.3 
 \\
GPT-4o  
&  26.1  & 65.1  & 19.5  & 64.9  & 8.6  & 53.1  & 17.0  & 57.8  & 9.7  & 51.9  & 16.9  & 59.6 
        \\ \hline
\multicolumn{13}{c}{\textit{\textbf{Compact LVLMs}}}  \\ \hline
Aquila-VL-2B &
3.2 & 44.6 & 2.6 & 46.7 & 0.4 & 43.1 & 1.9 & 41.5 & 2.3 & 44.1 & 2.9 & 44.2 

    \\
\rowcolor[HTML]{EFEFEF} Aquila-VL-2B* & 19.6 & 66.4 & 27.2 & 71.7 & 9.5 & 65.1 & 8.9 & 52.0 & 21.4 & 64.9 & 17.3 & 60.6 

     \\
InternVL2.5-2B  & 14.0 & 60.4 & 10.9 & 59.7 & 5.7 & 55.5 & 14.4 & 61.0 & 15.0 & 56.3 & 13.9 & 55.4 

     \\
\rowcolor[HTML]{EFEFEF}InternVL2.5-2B*  & 25.5 & 74.3 & 34.2 & 77.9 & 35.7 & 80.5 & 21.4 & 67.5 & 33.9 & 74.4 & 24.4 & 69.5 

  \\
DeepseekVL2-Tiny   & 3.2 & 49.9 & 3.4 & 51.2 & 1.3 & 54.2 & 3.8 & 51.7 & 1.9 & 49.3 & 5.3 & 49.0 

    \\
\rowcolor[HTML]{EFEFEF}DeepseekVL2-Tiny* & 24.2 & 75.4 & 40.6 & 84.5 & 33.6 & \textbf{83.4} & 20.8 & 71.7 & 30.0 & 76.5 & 29.7 & 75.0  \\

SmolVLM2-2.2B & 1.9 & 39.2  &  1.0 & 36.6  &  0.3 & 36.1  &  0.9 & 37.3 & 0.6 & 36.1 & 1.8 & 38.3 \\

\rowcolor[HTML]{EFEFEF}SmolVLM2-2.2B* 
& 11.2 & 53.2 &  10.8 & 52.7 & 9.1 & 55.0 & 8.9 & 51.1 & 13.9 & 48.0 & 15.7 & 58.6 \\

PaliGemma2-3B & 1.5 & 39.3 & 1.3 & 42.6 & 0.1 & 29.4 & 1.4 & 39.3 & 1.4 & 38.7 & 1.0 & 36.5 \\

\rowcolor[HTML]{EFEFEF}PaliGemma2-3B* & 18.0 & 63.1 & 13.5 & 54.0 & 8.7 & 51.3 & 13.3 & 57.1 & 12.3 & 49.7 & 14.7 & 57.4 \\

Qwen2.5-VL-3B  & 9.2 & 55.8 & 11.7 & 62.8 & 7.4 & 58.4 & 14.1 & 63.8 & 11.0 & 55.2 & 9.7 & 53.9 

   \\
\rowcolor[HTML]{EFEFEF}Qwen2.5-VL-3B*  & \textbf{29.3} & \textbf{77.8} & \textbf{46.5} & \textbf{86.2} & \textbf{42.7} & 83.0 & \textbf{28.6} & \textbf{73.6} & \textbf{42.5} & \textbf{86.5} & \textbf{31.1} & \textbf{76.1} 

     \\ \hline
\multicolumn{13}{c}
{\textit{\textbf{Cascade Pipelines}}}                  \\ \hline EasyOCR + LLM & 7.6 & 53.0 & 26.7 & 71.4 & 37.2 & 81.0 & 5.5 & 49.0 & 28.6 & 72.9 & 9.0 & 50.5 
\\ GOT-OCR + LLM & 23.2 & 71.9 & 33.1 & 77.5 & 39.3 & 81.4 & 13.8 & 59.9 & 32.8 & 76.2 & 21.5 & 63.0

\\ \bottomrule[1pt]
\end{tabular}
\caption{Detailed results for region-specific translation task (ZH-EN).}
\end{subtable}
\end{scriptsize}
\caption{Detailed evaluation results for region-specific translation task. Models marked with * indicate fine-tuning on our PATIMT train set. Best results are masked in \textbf{bold}. }
\label{figures:appendix_main_region}
\end{table*}
\begin{table*}[]
\centering
\begin{scriptsize}

\begin{subtable}{\textwidth}
\setlength{\tabcolsep}{0.4mm}
\renewcommand\arraystretch{1.3}
\begin{tabular}{lcccccccccccccccccc@{}}
\toprule[1pt]
& \multicolumn{3}{c}{\textbf{ads\&book\&poster}} & \multicolumn{3}{c}{\textbf{chart\&table}} & \multicolumn{3}{c}{\textbf{document}} & \multicolumn{3}{c}{\textbf{hand-written}} & \multicolumn{3}{c}{\textbf{infographics}} & \multicolumn{3}{c}{\textbf{natural\&street}} \\
\multirow{-3}{*}{\textbf{Model}} 
& 
\makebox[0.05\textwidth][c]{BLEU} & \makebox[0.04\textwidth][c]{COMET} & \makebox[0.04\textwidth][c]{IoU} & \makebox[0.05\textwidth][c]{BLEU} & \makebox[0.04\textwidth][c]{COMET} & \makebox[0.04\textwidth][c]{IoU} & 
\makebox[0.05\textwidth][c]{BLEU} & \makebox[0.04\textwidth][c]{COMET} & \makebox[0.04\textwidth][c]{IoU} & 
\makebox[0.05\textwidth][c]{BLEU} & \makebox[0.04\textwidth][c]{COMET} & \makebox[0.04\textwidth][c]{IoU} & 
\makebox[0.05\textwidth][c]{BLEU} & \makebox[0.04\textwidth][c]{COMET} & \makebox[0.04\textwidth][c]{IoU} & 
\makebox[0.05\textwidth][c]{BLEU} & \makebox[0.04\textwidth][c]{COMET} & \makebox[0.04\textwidth][c]{IoU} 

\\ \hline
\multicolumn{19}{c}{\textit{\textbf{Proprietary LVLMs}}}      \\ \hline

Qwen2.5-VL-72B &
19.5 & 65.9 & 0.274 & 9.8 & 46.6 & 0.110 & 3.0 & 37.7 & 0.056 & 12.7 & 37.8 & 0.153 & 1.7 & 35.7 & 0.036 & 32.8 & 68.4 & 0.480

  \\
GPT-4o  
&   12.1 & 56.6 & 0.138 & 6.3 & 48.8 & 0.059 & 2.1 & 40.3 & 0.057 & 2.3 & 39.4 & 0.050 & 3.1 & 43.9 & 0.058 & 15.7 & 55.2 & 0.045 
         \\ \hline
\multicolumn{19}{c}{\textit{\textbf{Compact LVLMs}}}  \\ \hline
Aquila-VL-2B & 1.0 & 20.9 & 0.062 & 0.0 & 10.8 & 0.001 & 0.0 & 5.0 & 0.004 & 0.0 & 11.0 & 0.058 & 0.0 & 17.8 & 0.005 & 5.1 & 28.5 & 0.090

   \\
\rowcolor[HTML]{EFEFEF} Aquila-VL-2B* &  19.8 & 67.2 & 0.422 & 9.7 & 55.7 & 0.139 & 15.2 & 61.2 & 0.355 & 29.6 & 77.4 & 0.679 & 14.4 & 59.6 & 0.210 & 28.9 & 68.9 & 0.347 

  \\
InternVL2.5-2B                        & 8.1 & 53.7 & 0.056 & 2.2 & 43.7 & 0.015 & 1.6 & 40.8 & 0.035 & 9.3 & 51.1 & 0.185 & 1.6 & 41.8 & 0.030 & 16.6 & 53.8 & 0.022 
 
    \\
\rowcolor[HTML]{EFEFEF}InternVL2.5-2B*  & 28.8 & 72.7 & \textbf{0.512} & 10.6 & 47.4 & 0.212 & 14.1 & 52.8 & 0.428 & 25.3 & 68.6 & 0.641 & 9.2 & 45.1 & 0.246 & 32.2 & 70.9 & 0.428 

 \\
DeepseekVL2-Tiny   & 0.0 & 0.0 & 0.000 & 0.0 & 0.0 & 0.000 & 0.0 & 0.0 & 0.000 & 0.0 & 0.0 & 0.000 & 0.0 & 0.0 & 0.000 & 0.0 & 0.0 & 0.000 

 \\
\rowcolor[HTML]{EFEFEF}DeepseekVL2-Tiny* & 21.9 & 66.3 & 0.364 & 6.8 & 51.9 & 0.076 & 4.1 & 49.4 & 0.085 & 21.5 & 70.8 & 0.109 & 12.1 & 58.5 & 0.182 & 31.3 & 69.2 & 0.376   \\

SmolVLM2-2.2B & 0.0 & 0.0 & 0.000 & 0.0 & 0.0 & 0.000 & 0.0 & 0.0 & 0.000 & 0.0 & 0.0 & 0.000 & 0.0 & 0.0 & 0.000 & 0.0 & 0.0 & 0.000 \\

\rowcolor[HTML]{EFEFEF}SmolVLM2-2.2B* 
& 17.7 & 67.3 & 0.446 & 6.0 & 51.4 & 0.077 & 9.9 & 53.5 & 0.191 & 12.3 & 57.1 & 0.471 & 6.6 & 51.3 & 0.098 & 21.2 & 64.5 & 0.256 \\

PaliGemma2-3B & 0.0 & 0.0 & 0.000 & 0.0 & 0.0 & 0.000 & 0.0 & 0.0 & 0.000 & 0.0 & 0.0 & 0.000 & 0.0 & 0.0 & 0.000 & 0.0 & 0.0 & 0.000 \\

\rowcolor[HTML]{EFEFEF}PaliGemma2-3B* & 22.0 & 57.5 & 0.143 & 9.2 & 51.0 & 0.041 & 13.9 & 52.0 & 0.206 & 9.5 & 48.9 & 0.141 & 11.0 & 53.1 & 0.063 & 22.8 & 64.8 & 0.043 \\

Qwen2.5-VL-3B  &  5.3 & 19.0 & 0.093 & 9.8 & 46.6 & 0.110 & 1.2 & 15.9 & 0.071 & 0.0 & 18.1 & 0.084 & 3.5 & 16.2 & 0.059 & 0.1 & 1.7 & 0.023 
 \\
\rowcolor[HTML]{EFEFEF}Qwen2.5-VL-3B*  & \textbf{26.4} & \textbf{69.8} & 0.439 & \textbf{15.0} & \textbf{57.0} & \textbf{0.279} & \textbf{23.3} & \textbf{64.9} & \textbf{0.539} & \textbf{36.4} & \textbf{79.9} & \textbf{0.669} & \textbf{20.9} & \textbf{60.4} & \textbf{0.366} & \textbf{36.2} & \textbf{69.8} & \textbf{0.448}

 \\ \hline
\multicolumn{19}{c}
{\textit{\textbf{Cascade Pipelines}}}              \\ \hline EasyOCR + LLM & 11.0 & 54.9 & 0.312 & 5.7 & 52.7 & 0.251 & 1.2 & 42.3 & 0.094 & 0.3 & 29.0 & 0.087 & 3.1 & 49.5 & 0.222 & 11.9 & 53.7 & 0.372 \\ GOT-OCR + LLM & 6.9 & 41.3 & - & 2.3 & 35.7 & - & 14.6 & 48.2 & - & 38.6 & 75.6 & - & 2.9 & 34.3 & - & 4.1 & 40.3 & -

\\ \bottomrule[1pt]
\end{tabular}
\caption{Detailed results for full-image translation with grounding task (EN-ZH).}
\vspace{2mm}
\end{subtable}
\\

\begin{subtable}{\textwidth}
\setlength{\tabcolsep}{0.4mm}
\renewcommand\arraystretch{1.3}
\begin{tabular}{lcccccccccccccccccc@{}}
\toprule[1pt]
& \multicolumn{3}{c}{\textbf{ads\&book\&poster}} & \multicolumn{3}{c}{\textbf{chart\&table}} & \multicolumn{3}{c}{\textbf{document}} & \multicolumn{3}{c}{\textbf{hand-written}} & \multicolumn{3}{c}{\textbf{infographics}} & \multicolumn{3}{c}{\textbf{natural\&street}} \\
\multirow{-3}{*}{\textbf{Model}}     & \makebox[0.05\textwidth][c]{BLEU} & \makebox[0.04\textwidth][c]{COMET} & \makebox[0.04\textwidth][c]{IoU} & \makebox[0.05\textwidth][c]{BLEU} & \makebox[0.04\textwidth][c]{COMET} & \makebox[0.04\textwidth][c]{IoU} & 
\makebox[0.05\textwidth][c]{BLEU} & \makebox[0.04\textwidth][c]{COMET} & \makebox[0.04\textwidth][c]{IoU} & 
\makebox[0.05\textwidth][c]{BLEU} & \makebox[0.04\textwidth][c]{COMET} & \makebox[0.04\textwidth][c]{IoU} & 
\makebox[0.05\textwidth][c]{BLEU} & \makebox[0.04\textwidth][c]{COMET} & \makebox[0.04\textwidth][c]{IoU} & 
\makebox[0.05\textwidth][c]{BLEU} & \makebox[0.04\textwidth][c]{COMET} & \makebox[0.04\textwidth][c]{IoU} 
\\ \hline
\multicolumn{19}{c}{\textit{\textbf{Proprietary LVLMs}}}      \\ \hline

Qwen2.5-VL-72B &
19.5 & 66.1 & 0.51 & 9.0 & 47.6 & 0.053 & 4.2 & 43.1 & 0.034 & 16.0 & 59.5 & 0.409 & 2.0 & 39.9 & 0.058 & 16.8 & 62.0 & 0.439

  \\
GPT-4o  
&   16.2 & 54.7 & 0.149 & 8.7 & 53.5 & 0.059 & 3.0 & 47.2 & 0.078 & 8.1 & 42.3 & 0.105 & 3.0 & 47.1 & 0.086 & 10.1 & 46.0 & 0.089
       \\ \hline
\multicolumn{19}{c}{\textit{\textbf{Compact LVLMs}}}  \\ \hline
Aquila-VL-2B & 
0.6 & 23.0 & 0.069 & 0.3 & 21.4 & 0.007 & 0.0 & 12.9 & 0.014 & 0.2 & 24.5 & 0.144 & 0.1 & 17.9 & 0.008 & 0.4 & 25.2 & 0.094

   \\
\rowcolor[HTML]{EFEFEF} Aquila-VL-2B* & 10.6 & 56.6 & 0.419 & 7.5 & 54.3 & 0.124 & 5.5 & 54.7 & 0.339 & 7.8 & 48.9 & 0.482 & 4.8 & 51.0 & 0.247 & 8.4 & 53.0 & 0.379

  \\
InternVL2.5-2B                        & 7.6 & 51.1 & 0.085 & 2.9 & 46.4 & 0.021 & 3.2 & 49.5 & 0.037 & 9.0 & 48.0 & 0.064 & 1.7 & 43.0 & 0.042 & 7.2 & 45.4 & 0.032

    \\
\rowcolor[HTML]{EFEFEF}InternVL2.5-2B*  & 13.1 & 63.3 & 0.487 & 7.6 & 51.2 & 0.21 & 4.1 & 42.0 & 0.315 & 16.0 & 62.6 & 0.669 & 8.5 & 46.0 & 0.346 & 15.9 & \textbf{62.4} & 0.53

 \\
DeepseekVL2-Tiny   & 0.0 & 0.0 & 0.000 & 0.0 & 0.0 & 0.000 & 0.0 & 0.0 & 0.000 & 0.0 & 0.0 & 0.000 & 0.0 & 0.0 & 0.000 & 0.0 & 0.0 & 0.000

 \\
\rowcolor[HTML]{EFEFEF}DeepseekVL2-Tiny* & 13.5 & 59.6 & 0.393 & 9.4 & 56.8 & 0.139 & 11.7 & 56.8 & 0.197 & 13.8 & 60.0 & 0.583 & 5.1 & 51.0 & 0.153 & 13.8 & 59.4 & 0.414
  \\

SmolVLM2-2.2B & 0.0 & 0.0 & 0.000 & 0.0 & 0.0 & 0.000 & 0.0 & 0.0 & 0.000 & 0.0 & 0.0 & 0.000 & 0.0 & 0.0 & 0.000 & 0.0 & 0.0 & 0.000 \\

\rowcolor[HTML]{EFEFEF}SmolVLM2-2.2B* 
& 9.2 & 50.3 & 0.274 & 9.2 & 49.8 & 0.143 & 8.4 & 48.9 & 0.080 & 7.3 & 47.2 & 0.513 & 10.5 & 52.7 & 0.107 & 16.8 & 58.5 & 0.292 \\

PaliGemma2-3B & 0.0 & 0.0 & 0.000 & 0.0 & 0.0 & 0.000 & 0.0 & 0.0 & 0.000 & 0.0 & 0.0 & 0.000 & 0.0 & 0.0 & 0.000 & 0.0 & 0.0 & 0.000 \\

\rowcolor[HTML]{EFEFEF}PaliGemma2-3B* & 9.1 & 52.3 & 0.145 & 9.7 & 48.7 & 0.106 & 15.7 & 61.3 & 0.088 & 8.1 & 47.5 & 0.208 & 9.0 & 49.9 & 0.095 & 13.7 & 50.7 & 0.301 \\

Qwen2.5-VL-3B  & 3.0 & 18.5 & 0.117 & 2.9 & 28.9 & 0.023 & 1.0 & 15.4 & 0.022 & 3.6 & 22.6 & 0.152 & 0.9 & 11.5 & 0.035 & 2.0 & 9.8 & 0.061
  \\

\rowcolor[HTML]{EFEFEF}Qwen2.5-VL-3B* & \textbf{15.9} & \textbf{64.5} & \textbf{0.481} & \textbf{15.8} & 58.5 & 0.236 & \textbf{21.3} & 56.7 & \textbf{0.381} & \textbf{22.6} & \textbf{63.1} & \textbf{0.62} & \textbf{15.7} & \textbf{54.4} & \textbf{0.389} & \textbf{14.0} & 59.0 & \textbf{0.452}

 \\ \hline
\multicolumn{19}{c}
{\textit{\textbf{Cascade Pipelines}}}              \\ \hline EasyOCR + LLM & 7.3 & 52.5 & 0.44 & 11.3 & \textbf{60.9} & \textbf{0.332} & 2.9 & 50.4 & 0.125 & 5.0 & 40.6 & 0.382 & 2.2 & 46.6 & 0.195 & 3.9 & 43.1 & 0.357 \\ GOT-OCR + LLM & 3.6 & 47.0 & - & 3.8 & 40.5 & - & 16.7 & \textbf{61.5} & - & 5.2 & 48.4 & - & 5.6 & 42.7 & - & 3.1 & 44.1 & -

\\ \bottomrule[1pt]
\end{tabular}
\caption{Detailed results for full-image translation with grounding task (ZH-EN).}
\vspace{2mm}
\end{subtable}
\end{scriptsize}
\caption{Detailed evaluation results for full-image translation with grounding task. Models marked with * indicate fine-tuning on our PATIMT train set. Best results are masked in \textbf{bold}.}
\label{figures:appendix_main_full}
\end{table*}

This section presents the complete results of our main experiments. Table \ref{figures:appendix_main_region} reports the results for the region-specific translation task, while Table \ref{figures:appendix_main_full} provides detailed results for the full-image translation with grounding task.
From these tables, we observe that Qwen2.5-VL-3B achieves the best performance across most metrics after fine-tuning. Additionally, models such as Aquila-VL-2B and DeepseekVL2-Tiny demonstrate strong performance despite their relatively limited foundational capabilities. 

Moerover, our evaluation reveals distinct performance patterns across different domains:
\begin{itemize}
\begin{CJK}{UTF8}{gbsn}
    \setlength{\itemsep}{0.1cm}
    \setlength{\parsep}{1pt}
    \setlength{\parskip}{1pt}
    \item \textbf{Easy Domains}.\quad Most LVLMs achieve high performance in easy domains such as ads\&books\&posters, and natural scenes\&street view. The improvement is limited in these domains because the number of text regions is usually small and often dominates the image, making them easier to recognize. In contrast, domains like charts\&tables and hand-written text show significant improvement. Charts\&tables contain small characters, while hand-written text contains characters that are harder to recognize.
    \item \textbf{Hard Domains}.\quad Performance in document and infographic domains is similar across all LVLMs. Both domains contain long paragraphs and small characters. The primary difference lies in layout: documents typically have a structured layout, while infographics have a more random layout. However, experiments show that this difference does not significantly impact performance. We attribute this to the models' ability to accurately locate texts after fine-tuning across multiple domains.
    \item \textbf{However, we observe the opposite pattern in ZH-EN full-image translation with grounding, where models perform better on hard domains than on easy domains}.\quad We attribute this to the fact that difficult scenarios typically contain longer, semantically coherent text, which provides richer contextual information to guide translation. In contrast, simple scenarios often feature short, context-deficient phrases, such as advertising slogans or highly localized Chinese expressions, that most models are poorly trained to handle (e.g. 新华书店(Xinhua Bookstore) is translated to New China Bookstore). Even when these phrases are accurately recognized, they show suboptimal translation quality for such culturally embedded phrases.
\end{CJK}
\end{itemize}

\end{document}